%% file: main.tex
\documentclass{article} 
\usepackage{iclr2027_conference,times}

\input{preamble_common}

\arxivtrue 
\iclrpreprintcopy

\begin{document}

\maketitle

\begin{abstract}
Concept bottleneck models (CBMs) make predictions inspectable and intervenable by routing them through human-interpretable concepts, but originally required concept annotations.
Annotation-free variants remove this requirement, but typically use large concept vocabularies, static at both training and inference, producing bottlenecks larger than any task or prediction needs and harder to inspect.
We propose \method, a single recipe for building compact, task-specific CBMs.
\method assumes only a frozen backbone and an unsupervised concept pool, such as a sparse autoencoder. A black-box teacher trained on the same backbone then guides selection: each round adds the concept that best explains the bottleneck's current failures, with candidates restricted to regions of teacher/student disagreement.
Trained with nested dropout over this selection order, the final bottleneck predicts accurately from any concept prefix, so inference spends fewer concepts on inputs it is confident about early and more on hard ones.
Since no stage is modality-specific, a new domain and task requires swapping only the backbone and concept pool.
Across six vision and two text datasets and two teacher paradigms, \method yields bottlenecks with fewer concepts and higher concept consistency than annotation-free baselines, while staying close to the black-box reference.
Overall, \method shows that going annotation-free need not mean going large: a small, discovered vocabulary can be as expressive as a much larger, fixed one.

\end{abstract}

\input{content/sections/1_intro}

\input{content/sections/2_related}

\input{content/sections/3_method}

\input{content/sections/4_evaluations}

\input{content/sections/5_discussion}

\input{content/sections/6_conclusion}


\input{content/sections/7_reproducibility}

\input{content/sections/8_ai_use_statement}



\bibliographystyle{plainnat}
\bibliography{main_bib}


\input{content/sections/9_supplementary}

\end{document}

%% file: preamble_common.tex
\usepackage[utf8]{inputenc} 
\usepackage[T1]{fontenc}    
\usepackage{url}            
\usepackage{booktabs}       
\usepackage{amsfonts}       
\usepackage{nicefrac}       
\usepackage{microtype}      
\usepackage{amsmath}
\usepackage{graphicx}
\usepackage[colorlinks=true, allcolors=blue]{hyperref}
\usepackage[dvipsnames,table,xcdraw]{xcolor}
\usepackage{xspace}
\usepackage{multicol}
\usepackage{multirow}
\usepackage{booktabs} 
\usepackage{makecell} 
\usepackage{wrapfig}
\usepackage{graphicx} 
\usepackage{algorithm}
\usepackage{algpseudocode}
\usepackage{caption}
\usepackage{arydshln} 
\usepackage{tikz}
\usetikzlibrary{patterns}
\usetikzlibrary{plotmarks}
\usepackage{pgfplots}
\usepackage{listings}
\usepackage{pgfplotstable}
\usepackage{placeins}
\pgfplotsset{compat=1.18}

\usepackage{algorithm}
\usepackage{algpseudocode}



\newcommand{\method}{\textsc{AnyBottle}\xspace}
\newcommand{\methodfull}{\textsc{AnyBottle}\xspace}

\DeclareMathOperator*{\argmax}{arg\,max}

\usepackage[textwidth=1.5cm]{todonotes}
\newcounter{mycomment}

\title{AnyBottle: A Recipe to Only Keep the \\Concepts You Really Need}

\author{
\textbf{Wolfgang Stammer$^{1}$,
Sukrut Rao$^{1}$,
Hevra Petekkaya$^{1}$,
David Steinmann$^{2,3}$,
Bernt Schiele$^{1}$}\\[0.6em]
$^{1}$Max Planck Institute for Informatics, Saarland Informatics Campus\\ 
$^{2}$Technical University of Darmstadt
$^{3}$Hessian.AI
}

\newif\ifarxiv
\arxivfalse

%% file: content/sections/1_intro.tex
\section{Introduction}

Concept bottleneck models (CBMs) provide interpretable decompositions of decisions across domains: a feature extractor maps inputs to human-interpretable concepts, combined via a simple (often linear) predictor. While originally requiring annotated concepts \citep{koh2020concept}, recent approaches mainly use CLIP's language alignment, either via CLIP-based supervision \citep{oikarinen2023label,yang2023language} or unsupervised concept discovery on a CLIP feature extractor \citep{rao2024discover,bhalla2024interpreting,wittenmayer2026cfm}, scaling in a task-agnostic manner without concept-level annotations.

\begin{figure}[t]
    \centering
    \includegraphics[width=0.9\linewidth]{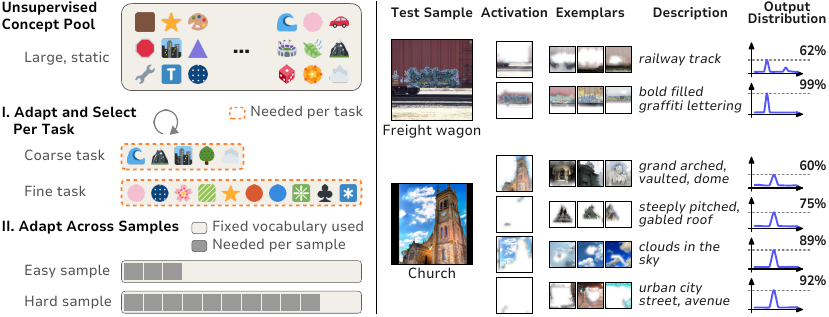}
    \caption{\textbf{Only as many concepts as needed.} \textit{Left:} existing annotation-free concept bottlenecks draw from a large, static concept pool, using far more concepts than a given task needs, and the same fixed vocabulary for every sample regardless of difficulty. \method instead selects a concept set sized to the task, and lets inference use only as many of those concepts as each sample actually requires. \textit{Right:} two test samples illustrating this sample-level adaptivity. 
    }
    \label{fig:motivation}
\end{figure}

Despite their utility, large, overcomplete concept vocabularies of thousands of candidates hurt interpretability on specialized tasks, where far fewer concepts often suffice. Globally, having many influential concepts causes duplication and feature splitting, complicating debugging and intervention; locally, individual predictions may use more concepts than necessary (\autoref{fig:motivation}, left). Reliance on a CLIP-like backbone \citep{oikarinen2023label, bhalla2024interpreting, shang2024incremental, tan2024explain} and web-scale concept extractors further limits applicability, as these rarely cover a target domain's fine-grained or specialized concepts.

To address this, we propose \method, a general recipe for compact, task-specific CBMs that keep only the concepts each output and sample needs. We start from an opaque teacher predictor and an initial, large, task-agnostic concept pool built with a sparse autoencoder \citep{bricken_towards_2023}. Keeping feature extractor and teacher frozen, \method selects concepts iteratively: each round, we identify the samples with the highest teacher error and add concepts covering parts of these inputs that the teacher uses, but the current concepts do not, yielding a compact concept set that stays task-relevant. The classifier is trained via nested dropout, enabling prediction from any number of concepts and letting inference spend fewer concepts on easy inputs and more on hard ones (\autoref{fig:motivation}, right). This small concept space also enables automatic, agentic naming \citep{shaham2025maia}, testing concept meaning via targeted interventions rather than similarity to a fixed CLIP vocabulary \citep{bhalla2024interpreting}. Finally, since selection assumes only a frozen backbone and an unsupervised concept pool, the recipe transfers across domains and modalities by swapping just these two components. Across six vision and two text datasets, \method matches or exceeds teacher accuracy using an order of magnitude fewer, and individually more coherent, concepts than annotation-free baselines, resulting in a bottleneck only as large and detailed, as the task and sample demand.

In summary, our contributions are threefold: (i) a teacher-guided concept selection procedure that reduces disagreement with an opaque teacher to build a compact, task-specific bottleneck; (ii) a nested-dropout predictor that exploits this ordering to predict accurately from any concept prefix, spending fewer concepts on confident inputs; and (iii) \method itself, a recipe requiring no human concept annotation or language-aligned backbone, which we show transfers across domains and modalities on vision and text.

%% file: content/sections/2_related.tex
\section{Related Work}

\textbf{Concept Bottleneck Models (CBMs).} 
CBMs \citep{koh2020concept, StammerSK21, knab2026s} mark an important moment in the growing interest in concept-based models \citep{YehKR21,fel2023holistic, poeta2023concept}, promising interpretable predictions and a structured interface for interaction. While the original framework relied on fully supervised concept annotations, subsequent research has relaxed this via unsupervised concepts \citep{sawada2022concept}, leveraging pretrained vision-language models like CLIP for concept extraction \citep{bhalla2024interpreting, yang2023language, oikarinen2023label, panousis2024coarse, yamaguchi2025zeroshot}, or employing fully unsupervised concept discovery \citep{stammer2024neural, rao2024discover, schrodi2024concept, Schut25chess}. Most approaches produce a fixed concept layer, set once and used as-is. \method instead builds its bottleneck dynamically, keeping only the concepts a teacher's behavior calls for.

\textbf{Compact and Sparse Concept Selection.} A separate line targets compactness directly: how few concepts suffice to solve a task. \citet{chattopadhyay2023information} cast Orthogonal Matching Pursuit as Information Pursuit (CLIP-IP-OMP), greedily selecting CLIP text-embedding atoms by residual correlation, later extended to a hierarchical dictionary \citep{nguyen2026hierarchical}. Res-CBM \citep{shang2024incremental} completes a CLIP-text bank with optimizable vectors where existing concepts under-explain a class. OpenCBM \citep{tan2024explain} instead reconstructs an already-trained classifier post hoc via greedy nearest-concept-to-residual steps. Sparse-CBM \citep{semenov2024sparse} and UCBM \citep{schrodi2024concept} pursue compactness via Gumbel-softmax and input-dependent gating instead. In text, TBM \citep{ludan2023interpretable} iteratively prompts an LLM for concepts from misclassified examples, while CT-CBM \citep{bhan2025towards} scores unsupervised candidates by activation importance until a residual branch plateaus. Most select from a language-aligned (CLIP-text or LLM-generated) bank; \method instead selects greedily by residual over an unsupervised SAE dictionary, filtering candidates by teacher-student attribution disagreement.

\textbf{Adaptive, Variable-Budget Inference.}
A separate strand asks how many concepts a \emph{given input} needs, and whether one model can serve multiple budgets without retraining, an idea recurring across ordered, prefix-usable representations \citep{bachmann2025flextok, wen2025principal}. Nested Dropout \citep{rippel2014learning} stochastically drops an ordered suffix of hidden units so any prefix remains usable, an effect Matryoshka Representation Learning \citep{kusupati2022matryoshka} achieves via nested loss terms instead; Matryoshka CBM \citep{chen2026matryoshka} applies this to concept bottlenecks, ordering a given vocabulary and serving fixed budgets via nested heads. The Variational Information Pursuit line \citep{chattopadhyayvariational, chattopadhyay2024bootstrapping} instead studies confidence-driven sequential querying, but over a user-defined or LLM-generated query set. \method instead applies Nested Dropout to its own teacher-discovered vocabulary, letting confidence decide how many concepts to consume.

\textbf{Sparse Autoencoders as a Concept Source.}
SAEs \citep{bricken_towards_2023, templeton_scaling_2024, bussmann2025Matryoshka} are increasingly used as a concept source in place of a fixed, LLM/VLM-queried vocabulary \citep{oikarinen2023label, shang2024incremental}. \citet{rao2024discover} discovers concepts via SAE first and names them afterward; \citet{wittenmayer2026cfm} pursues a related decomposition at foundation-model scale. Concept-SAE \citep{ding2025conceptsae} and VL-SAE \citep{shen2025vlsae} instead inject supervision into the SAE objective, aligning a subset of latents to externally specified concepts. \citet{rocchihenry2025geometric} argue CBMs and SAEs are the same geometric object, differing only in how their cone of directions is selected, supervision versus sparse coding; \method sits between these poles, performing teacher-supervised selection over an already-trained, frozen SAE dictionary.

%% file: content/sections/3_method.tex
\section{AnyBottle}
\label{sec:method}

\begin{figure}[t]
    \centering
    \includegraphics[width=0.95\textwidth]{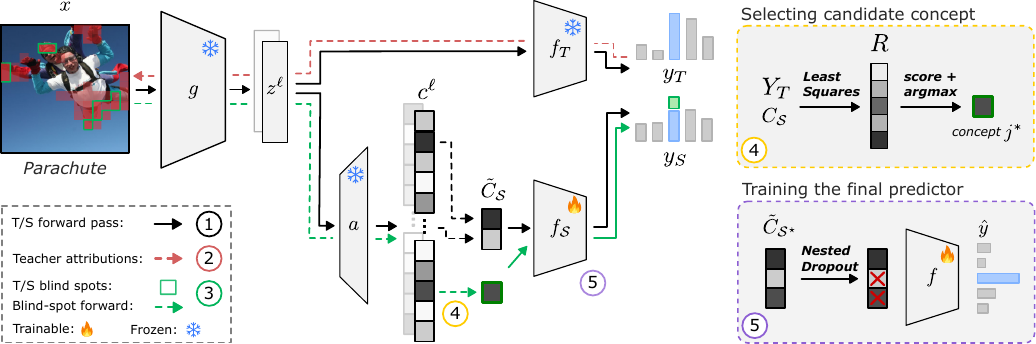}
    \caption{\textbf{Overview of \method.} An input $x$ is encoded by the frozen backbone $g$ into token embeddings $z^\ell$, which a sparse autoencoder $a$ maps to concept activations $c^\ell$. Each selection round repeats four steps: (1) a forward pass through the teacher $f_T$ and current student $f_\mathcal{S}$; (2) teacher attributions over the input; (3) the resulting teacher/student blind-spots (green), used to restrict the SAE to blind-spot tokens; and (4) scoring the resulting candidates by least-squares residual $R$ against the teacher's output $Y_T$ to pick the next concept $j^\star$, added to $\mathcal{S}$. This repeats until selection converges on $\mathcal{S}^\star$. Once selection terminates, (5) the final predictor $f$ is trained once on rescaled $\tilde{C}_{\mathcal{S}^\star}$ with nested dropout, yielding a head that predicts accurately from any concept prefix.}
    \label{fig:main}
\end{figure}

In this section, we describe \method's process for obtaining a compact, task-specific CBM from an unsupervised concept pool. It follows the bottom-up approach of performing concept grounding before semantics \citep{knab2026s}. An SAE first grounds a large pool of unnamed patterns (\autoref{sec:method_pool}). \method then selects among them by their statistical relationship to a black-box teacher's behavior (\autoref{sec:method_selection}). The predictor is trained on the final pool with nested dropout for a sample-adaptive concept budget at inference (\autoref{sec:method_nested}). Naming the selected concepts happens last, after selection is complete (\autoref{sec:method_naming}). \autoref{fig:main} overviews the recipe (pseudocode: \autoref{alg:anybottle}); details follow.

\textbf{Notation.} We adopt the formalism of \citet{knab2026s}: a CBM is an input encoder $g: \mathcal{X} \rightarrow \mathcal{Z}$, a concept grounding mechanism $h: \mathcal{Z} \rightarrow \mathcal{C}$ mapping embeddings to concept activations, and a predictor $f: \mathcal{C} \rightarrow \mathcal{Y}$, trained on $\{(\mathbf{x}_i, \mathbf{y}_i)\}_{i=1}^N \subseteq (\mathcal{X}, \mathcal{Y})$ with $\mathbf{y}_i$ a class label or regression target. Encoders often operate at the token level (e.g.\ image patches, word tokens) rather than on a single global embedding: we write $\mathbf{z}^\ell \in \mathbb{R}^D$, $\ell \in \{1, \dots, L\}$, for an individual token's embedding, and $\mathbf{z} = \frac{1}{L}\sum_{\ell=1}^L \mathbf{z}^\ell$ for the pooled, global embedding.

\method follows a teacher-student approach. The teacher is a black-box predictor, e.g.\ an MLP head, trained on the frozen encoder's global embeddings: $f_T(\mathbf{z}) = \mathbf{y}_T \in \mathbb{R}^{|\mathcal{Y}|}$. \method's predictor $f$ is trained on the activations $C_{\mathcal{S}^\star}$ of a small, selected subset of concept indices $\mathcal{S}^\star \subseteq \{1, \ldots, M\}$, chosen so $f$ predicts $\mathbf{y}_T$ as closely as possible. Since $\mathcal{S}^\star$ is unknown in advance, selection proceeds in rounds: each round, the current predictor $f_\mathcal{S}$ is trained on the current candidate set $\mathcal{S}$ and guides which concept to add next (\autoref{sec:method_selection}). Once selection converges on $\mathcal{S}^\star$, $f$ is trained from $f_\mathcal{S}$ with nested dropout (\autoref{sec:method_nested}) into the final, sample-adaptive predictor.

\subsection{Obtaining the Initial Concept Pool}\label{sec:method_pool}
\method instantiates the concept space $\mathcal{C}$ with an unsupervised concept pool: a sparse autoencoder (SAE) $a: \mathcal{Z} \rightarrow \mathcal{C}$, trained on token embeddings from the frozen encoder $g$. Applied independently to each token, it gives candidate concept activations $\mathbf{c}^\ell = a(\mathbf{z}^\ell) \in \mathbb{R}^M_{\geq 0}$ for $\ell \in \{1, \dots, L\}$: the activations of $M$ candidate concepts, shared across all inputs. We first train the SAE on a general corpus, without task-specific information. Following \citet{braun2024identifying}, it can then be fine-tuned to the target task: we pass the SAE's reconstructed embedding $\hat{\mathbf{z}}$ through the frozen teacher head in place of $\mathbf{z}$, minimizing the KL divergence between $f_T(\mathbf{z})$ and $f_T(\hat{\mathbf{z}})$ jointly with a standard reconstruction term. In practice, this can make concepts more task-suitable and yield a more compact concept space.

For selection, we pool each concept's token-level activations into a single value per input: $c_{ij} = \max_\ell (c_{ij}^\ell)$, keeping concept $j$'s largest activation among all tokens of input $i$. We compute this once, over the full dataset, giving $C \in \mathbb{R}^{N \times M}$, and reuse it throughout selection.
Further note, SAE activations differ widely in strength across concepts, so we rescale before training any predictor. For each concept $j \in \mathcal{S}$, a per-concept scale $\pi_j$ is set from the $p$-th percentile of its training-set activations, giving $\tilde{C}_{ij} = \text{clip}\left(\frac{C_{ij}}{\pi_j + \epsilon}, 0, \tau_\text{max}\right)$.
This outlier-robust rescaling is refit whenever $\mathcal{S}$ changes; both $f_\mathcal{S}$ and the final $f$ train and evaluate only on $\tilde{C}$, never raw $C$.

\subsection{Adaptive Concept Selection}\label{sec:method_selection}
Given the pooled activation matrix $C$ and the teacher $f_T$, \method iteratively expands a concept selection $\mathcal{S}$, inspired by Orthogonal Matching Pursuit (OMP) \citep{pati1993orthogonal}. Each round adds the concept that best reduces the residual between the current selection and the teacher's output, then trains a predictor $f_\mathcal{S}$ on the new selection, which guides the next round. We write $\mathbf{c}_j \in \mathbb{R}^N$ for the $j$-th column of $C$: concept $j$'s pooled activation across all $N$ inputs.

Throughout this subsection, $C_\mathcal{S}$ (the columns of $C$ indexed by $\mathcal{S}$) and $\mathbf{c}_j$ denote these raw, pooled activations, not the rescaled $\tilde{C}$ of \autoref{sec:method_pool}. Both the residual fit (\autoref{eq:residual}) and candidate scoring (\autoref{eq:scoring}) instead use an unfit, column-wise normalization, $\|\mathbf{c}_j\|_2$, recomputed each round; this keeps candidate ranking scale-invariant, independent of the rescaling used to train the predictor. Selection is driven by sample- and region-level (attribution) disagreement between $f_T$ and $f_\mathcal{S}$, described below.

\textbf{Computing the Residual.}
We write $Y_T \in \mathbb{R}^{N \times |\mathcal{Y}|}$ for the teacher's outputs over the full dataset. Each round, we fit $C_\mathcal{S}$ to $Y_T$ via least squares, augmenting $C_\mathcal{S}$ with a constant column, $\hat{C}_\mathcal{S} = [C_\mathcal{S}, \mathbf{1}]$, so that $\mathcal{S}$ only needs to explain deviations from $Y_T$'s mean rather than the mean itself: without this column, a concept that fires near-uniformly across the dataset could score as informative purely by matching the residual's mean, rather than by explaining any per-example variance. This yields coefficients $\hat{W} \in \mathbb{R}^{(|\mathcal{S}|+1) \times |\mathcal{Y}|}$ and a residual $R$:
\begin{equation}\label{eq:residual}
    \hat{W} = (\hat{C}_\mathcal{S}^\top \hat{C}_\mathcal{S})^{-1} \hat{C}_\mathcal{S}^\top Y_T, \hspace{2cm} R = Y_T - \hat{C}_\mathcal{S} \hat{W}
\end{equation}
with $W$ recovered by dropping $\hat{W}$'s intercept row. $R$ then selects the next concept from $\{1, \dots, M\} \setminus \mathcal{S}$, choosing whichever best explains it.

\textbf{Candidate Proposal via Attribution Disagreement.} Scoring every concept against $R$ each round is expensive and unnecessary, since most are irrelevant to the current failure mode. \method instead restricts candidates using token-level disagreement between the teacher's ($f_T$) and current predictor's ($f_\mathcal{S}$) attributions. We take the $k$ inputs with largest residual norm $\|R_i\|_2$ as focus set $\mathcal{I}$, and for each $i \in \mathcal{I}$ compute a token-importance map for $f_T$ via gradient attribution w.r.t.\ the ground-truth label; for $f_\mathcal{S}$, attribution is computed analytically, each selected concept's contribution to the target-class logit (its head weight times pooled activation) is assigned to the token where the concept attains its maximum activation. Tokens the teacher attends to but $f_\mathcal{S}$ does not are this round's \emph{blind-spots} $B_i$; summing per-token activations over them, $\text{agg}_i = \sum_{\ell \in B_i} \mathbf{c}^\ell$, and dropping already-selected or below-threshold concepts leaves the $r$ highest-activating candidates per sample.

Aggregating across $\mathcal{I}$, each surviving concept $j$ accumulates a distinct-sample count $n_j$ and mean activation strength $\delta_j$; we discard concepts found in too few samples and rank the rest by $\delta_j \cdot \log(1+n_j)$, favoring strong, broadly implicated concepts over one-off spikes. The top $q$ form the shortlist $\mathcal{J}$.

\textbf{Candidate Scoring and Selection.}
Given the shortlist $\mathcal{J}$ and the residual $R$, each candidate $j \in \mathcal{J}$ is scored by
\begin{equation}\label{eq:scoring}
    s_j = \frac{\|\mathbf{c}_j^\top R\|_2}{\|\mathbf{c}_j\|_2 + \epsilon}, \hspace{2cm} j^\star = \argmax_j s_j.
\end{equation}
The top-scoring candidate is added to the pool: $\mathcal{S} \leftarrow \mathcal{S} \cup \{j^\star\}$.
Notably, the first round has no prior concept set to fit against the target, so the first concept is instead scored against the mean-centered teacher output (Eq.~\ref{eq:residual} with $\mathcal{S} = \emptyset$), over the full candidate set: $\mathcal{J} = \{1, \dots, M\}$. For many-class tasks, this round can instead be jump-started with one heuristically class-specific concept per class, before residual-driven selection proceeds as usual.

\textbf{Retraining and Stopping.} After adding a concept, $R$ is recomputed and $f_\mathcal{S}$ retrained on $\tilde{C}_\mathcal{S}$, providing validation accuracy for model selection and student attribution for the next round. We track explained variance $\text{EV}_\mathcal{S} = 1 - \text{var}(R)/\text{var}(Y_T)$ as a diagnostic: as $\mathcal{S}$ grows, $\text{EV}_\mathcal{S}$ approaches the upper bound of the SAE's dictionary, serving as stopping signal and concept-pool-quality diagnostic.

Selection terminates when: (1) no candidate survives attribution-based filtering, or all candidate scores fall below a minimum threshold; (2) the accuracy gap to the teacher falls within a tolerance; (3) neither validation accuracy nor $\text{EV}_\mathcal{S}$ improves for a set number of rounds; or (4) a round limit is reached. We then return the best-validation-accuracy concept set $\mathcal{S}^\star$, its predictor $f_{\mathcal{S}^\star}$, and its per-concept scales $\{\pi_j\}_{j \in \mathcal{S}^\star}$, retaining the discovery order for training the final predictor.

\subsection{Nested Dropout}\label{sec:method_nested}
Once selection terminates with $\mathcal{S}^\star$ and its discovery order fixed, we initialize a final predictor $f$ from $f_{\mathcal{S}^\star}$ and retrain it on $\tilde{C}_{\mathcal{S}^\star}$ using nested dropout \citep{rippel2014learning}. At each training step, we sample a prefix length $l$ uniformly from $\{1, \dots, |\mathcal{S}^\star|\}$, zero every concept activation beyond the $l$-th in discovery order, and train $f$ on the resulting $\tilde{C}^{(l)}_{\mathcal{S}^\star}$ with cross-entropy loss against ground-truth labels. Since every step truncates to a new sampled prefix, $f$ learns to predict correctly from any prefix length, not just the full set. This enables an optional adaptive inference mode: concepts are consumed one at a time in discovery order, and inference halts at the shortest prefix $l$ whose predicted class confidence (max softmax of $f(\tilde{C}^{(l)}_{\mathcal{S}^\star})$) exceeds a threshold, with concepts below a minimum per-input activation masked out to retain only those that genuinely fire. This adapts the concept budget per input: fewer concepts for confident inputs, more for harder ones.

\subsection{Concept Naming}\label{sec:method_naming}
Because \method\ selects only the concepts a task needs, naming applies to a handful rather than the full SAE dictionary. Related work commonly uses a single CLIP-similarity pass \citep{oikarinen2023label, yang2023language, prasse2025dcbm}, cheap but prone to missing a concept's exact counterexamples and boundaries. Recent agentic naming methods \citep{shaham2025maia} (cf. \autoref{sec:related_app}) instead test these boundaries directly, e.g.\ by generating or editing interventionist images to check whether a concept tracks color or shape; more tools and testing rounds improve reliability but are too costly over a full, unselected dictionary. Naming \method's few dozen selected concepts this way remains cheap enough as an automated first pass.
Even so, a human should still check these proposals, ideally interactively \citep{SchramowskiSTBH20, FriedrichSSK23b}: \method's compactness makes checking a few dozen names feasible where checking thousands is not. This also catches what automated naming alone cannot: a well-named concept the predictor may still rely on for the wrong reason, tracking a spurious correlation rather than the task itself; such shortcuts are easier to spot once the compact set is named and reviewed against domain knowledge \citep{steinmann2024navigating, geirhos2020shortcut, lapuschkin2019unmasking}.

%% file: content/sections/4_evaluations.tex
\section{Experimental Evaluations}\label{sec:eval}

We evaluate \method across multiple datasets and modalities to answer the following questions: does it match or exceed annotation-free CBM baselines on accuracy, compactness, and concept consistency (\textbf{RQ1}); does selection close genuine teacher-student disagreement, allocating more concepts to harder tasks (\textbf{RQ2}); does nested dropout let inference allocate more concepts to harder individual samples (\textbf{RQ3}); which components of the selection mechanism drive its compactness and quality (\textbf{RQ4}); and does the recipe transfer to text, across domains and teacher paradigms (\textbf{RQ5}).

\textbf{Datasets.} We evaluate on six vision and two text classification datasets, chosen to span coarse- and fine-grained discrimination, small and large label vocabularies, and general versus specialized domains. Vision: \emph{Imagenette}~\citep{howard2019imagenette} (10-class, coarse), \emph{Dogs-10} (10 fine-grained breeds), \emph{Machines-15} (15 fine-grained machine types) and \emph{Mixed-20} (20 animal and machine classes), \emph{Flowers-102}~\citep{nilsback2008automated} (102-class, fine-grained), and \emph{ISIC-7}~\citep{codella2019skin,tschandl2018ham10000} (7-class dermoscopic skin lesions, a specialised domain). Dogs-10, Machines-15, and Mixed-20 are curated class subsets of ImageNet; full class lists are given in Appendix~\ref{app:exp_setup}. Text: \emph{AG News}~\citep{zhang2015characterlevel} (topic classification) and \emph{PubMed-20k}~\citep{dernoncourt2017pubmed} (biomedical sentence-role classification). Further details are given in App.~\ref{app:exp_data_setup}.

\textbf{Models.} \method uses CLIP-DINOiser~\citep{wysoczanska2024clip} (vision) or Gemma \citep{team2024gemma} (text) as backbone, with concepts from a sparse autoencoder with task-aware finetuning. We compare against four annotation-free or compactness-oriented CBM baselines (Label-free CBM~\citep{oikarinen2023label}, Res-CBM~\citep{shang2024incremental}, DN-CBM~\citep{rao2024discover}, UCBM~\citep{schrodi2024concept}); backbones, concept sources, and classifier heads for each are given in App.~\ref{app:exp_model_setup}. For vision and the mlp text teacher, the black-box teacher is a two-layer MLP trained per-backbone, keeping reported gaps within one embedding space; the verbalizer teacher trains no parameters. 
We also compare against a linear probe over the full, unselected SAE dictionary, and one restricted to active neurons only, isolating selection gains from those of any smaller, arbitrary subset. Full configuration and hyperparameter details including automatic concept naming are in App.~\ref{app:exp_model_setup}.

\textbf{Metrics.} Our primary compactness measure is $K = |\mathcal{S}^\star|$, the selected concept set size (for full-dictionary SAE probes, the number of active SAE features). We report balanced accuracy throughout; since methods use different backbones, our primary metric is the accuracy gap $\Delta$ between student (or CBM baseline) and teacher (black-box model), on the same backbone. All results are averaged over three random seeds. Since $f$ predicts at any prefix length up to $K$, we evaluate across prefix lengths and report $\mathrm{C}@q$ for $q \in \{90, 95, 98\}$ (the smallest prefix reaching $q$ percent of the student's own full-budget accuracy), plus $\mathrm{C}@K = K$, the full discovered budget, for reference.
To assess concept quality independent of accuracy, we use the $\mathrm{C}^2$ score~\citep{parchami2026fact} for vision: semantic consistency of each concept's most-activating inputs, averaged over the concept set. \method's concepts are scored at the region level; the four CBM baselines produce only one scalar activation per image and are scored at the image level. For text, we report an autointerp score (AIS) \citep{bills2023language, paulo2024automatically}: an LLM describes a concept from its top-activating documents, a second LLM predicts its activation on held-out documents from that description alone, and the score is the correlation between predicted and true activation. Further details in App.~\ref{app:exp_metric_setup}.

\subsection{Experimental Evaluations}
\label{sec:exps}

\begin{figure}[t]
    \centering
    \includegraphics[width=\textwidth]{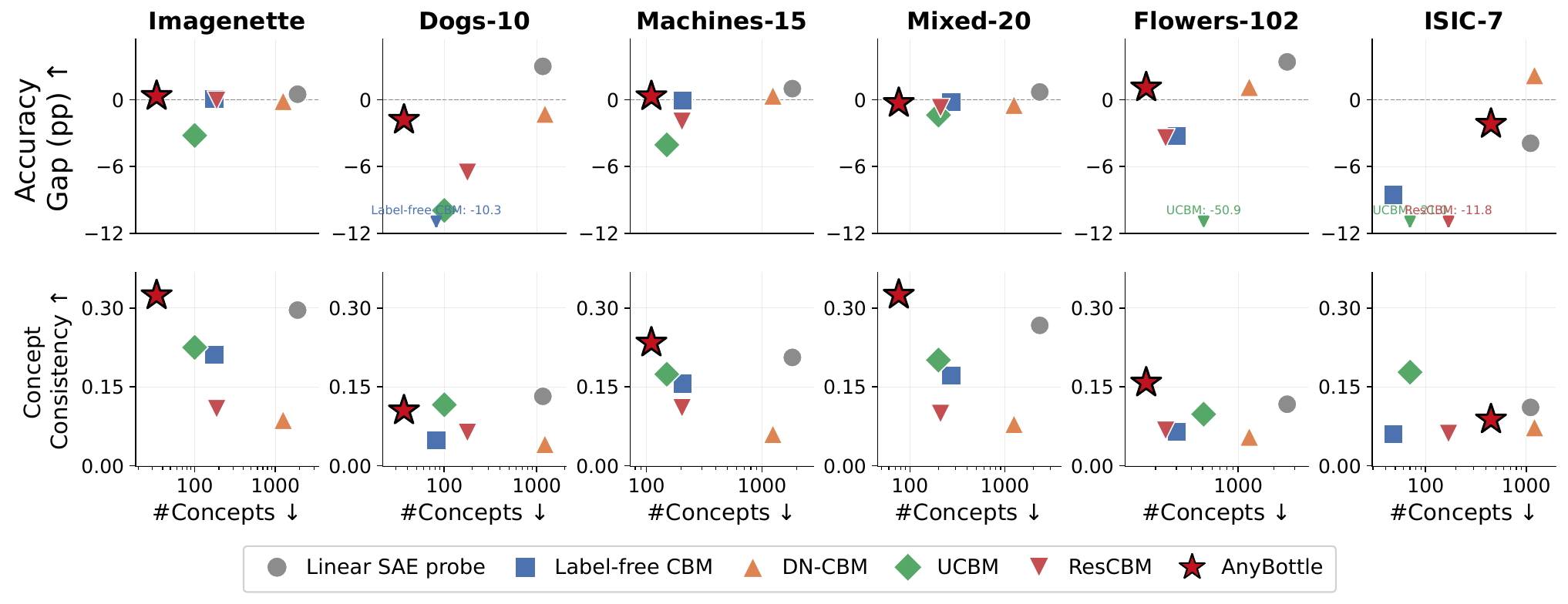}
    \caption{
    \textbf{Top: \method matches or exceeds black-box teacher accuracy at a fraction of the concepts used.} Accuracy gap ($\Delta$ (pp)) to the teacher, plotted against concept count (log scale); \method reaches the teacher across all six datasets with orders of magnitude fewer concepts than the baselines.
    \textbf{Bottom: \method's smaller bottleneck is also the more interpretable one.} Concept consistency ($C^2_\mathrm{img}$) at the same budgets matches or exceeds every baseline, showing the compactness above is not simply discarded capacity. Points are seed means.
    }
    \label{fig:main_results}
\end{figure}

\textbf{Adaptive Concept Bottlenecks on par with Black-box Models (RQ1).}
\autoref{fig:main_results} evaluates whether \method matches a black-box teacher's accuracy while using fewer, more consistent concepts than existing baselines across diverse vision tasks, without manual concept annotations.
\method reaches teacher accuracy on most datasets, comes within two points on Dogs-10 and ISIC-7, and surpasses it on Flowers-102: routing through a small, selected concept set instead of the teacher's embedding space costs almost no accuracy.
This compactness is not just discarded capacity: it also improves concept consistency. Against a linear probe over the same SAE dictionary, \method uses 16--60$\times$ fewer concepts on five of six datasets, while matching or improving image-level (and patch-level, cf. \autoref{fig:c2patch}) consistency on four. On ISIC-7, the one dataset needing a larger budget, \method's set is still only half the active SAE dictionary size and beats the probe's accuracy.
The pattern holds against the four annotation-free baselines: fewer concepts on all but one dataset, higher image-level consistency on all but two of twenty-four dataset--baseline pairs, and competitive accuracy throughout, with \method's teacher-gap $\Delta$ matching or beating every baseline except on ISIC-7, where DN-CBM's edge costs an order of magnitude more concepts. Raw accuracies stay broadly comparable across methods despite different backbones, except for a wider spread on ISIC-7 (\autoref{fig:main_clf_acc}).
Overall, \method matches or exceeds black-box teachers across tasks from coarse to fine-grained, at a fraction of the concepts annotation-free baselines require; ISIC-7, with its weak teacher, is the one setting where these gaps narrow across all methods.

\input{content/figures/cak_rounds}

\input{content/figures/adaptive_examples}

\textbf{\method Allocates Concepts Where Accuracy Demands Them (RQ2).}
\autoref{fig:cak} shows concepts needed by the nested-dropout head to reach $90\%$, $95\%$, and $98\%$ of its own full-budget accuracy, alongside the full discovered budget C@K. 
Most accuracy comes early: on Imagenette, Dogs-10, and Mixed-20, C@90 needs 27 concepts or fewer, with C@98 only a handful more. Machines-15 and Flowers-102 need a larger initial budget, matching their finer granularity, but show the same diminishing cost. 
ISIC-7 is the exception: C@90 already needs 135 concepts, and closing to C@98 needs more than double that, reflecting both task complexity (difficult and unbalanced classes) and a weak, noisy teacher. 
\input{content/tables/ablation_selection}
\autoref{fig:rounds_imagenette} (right) shows what drives this pattern round by round on Imagenette: round 11 adds a toothed-chain concept refining chainsaw recognition, round 15 a small-spheres concept improving golf-ball recognition, both closing blind spots the teacher attends to but the student does not. 
Early rounds close large, class-defining blind spots, while later rounds refine coverage with diminishing accuracy gain, confirming that allocation tracks task difficulty by closing genuine teacher-student disagreement, not by adding concepts uniformly.

\textbf{Adaptive Inference Allocates Concepts by Sample Difficulty (RQ3).}
\autoref{fig:adaptive_mixed} shows the nested-dropout head consuming concepts in discovery order at inference, for two easy samples and one hard sample. 
Both easy samples converge quickly: two concepts take a gorilla, and three a boxer, past 90\% confidence. The hard sample, an ice bear (Mixed-20), draws on far more of the budget, crossing 50\% confidence only after 5 concepts and needing 14 for 95\%. 
This is nested dropout's intended effect: a head accurate at any prefix length lets inference stop as soon as the prefix suffices, rather than always spending the full budget $K$.

\input{content/tables/text_cls}

\textbf{Ablating Selection Strategy (RQ4).}
To test \method's selection mechanism, \autoref{tab:ablation-selection} removes the prefiltering (PF), the localization step (Loc), and the least-squares (LS) scoring in turn. Removing any part progressively reduces compactness and concept consistency while increasing the accuracy gap to the teacher, confirming that each stage contributes to finding compact, consistent and accurate concept sets. Per-dataset results in \autoref{tab:ablation-selection-per-dataset}.
\autoref{tab:representation_taskaware} in the appendix reports two additional ablations. Continuous versus discrete concept representations: \method works with both types, though a discrete set usually has to be larger for the same accuracy. Additionally, task-aware SAE fine-tuning helps most when the pretrained dictionary is poorly matched to the target domain, notably on ISIC-7 and Flowers-102.

\textbf{Generalization Across Domains (RQ5).}
\autoref{tab:main_results_text} tests whether \method transfers beyond vision by swapping only the backbone and concept pool, along two axes: domain (general-topic AG News vs.\ specialized biomedical PubMed-20k) and teacher paradigm (mlp-clf, an MLP head on a frozen LM's hidden state, vs.\ verb-clf, a verbalizer teacher reading predictions directly off a frozen, instruction-tuned model's next-token distribution), each with its own backbones.
Against mlp-clf, \method closely matches teacher accuracy on both datasets and matches or outperforms the linear SAE probe in both accuracy and interpretability, using orders of magnitude fewer concepts. This mirrors the vision results: a small, selected concept set recovers near-teacher accuracy regardless of domain.
The verb-clf performs worse overall, only slightly above chance on PubMed-20k. \method again matches the teacher on AG News and substantially outperforms it on PubMed-20k, but falls short of the linear SAE probe there: teacher quality caps what \method's selection can recover. This confirms \method generalizes across teacher paradigms, while showing its limits when paired with a weak teacher.

%% file: content/figures/cak_rounds.tex
\begin{figure}[t]
\centering
\begin{minipage}[b]{0.265\textwidth}
\centering
\includegraphics[width=0.95\linewidth]{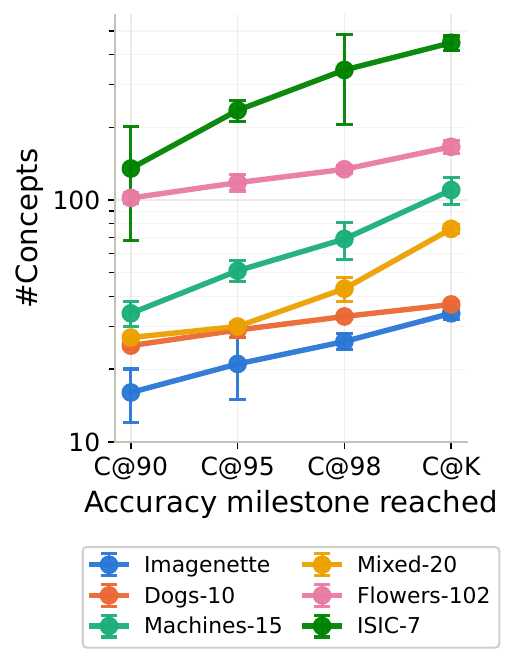}
\end{minipage}
\hfill
\begin{minipage}[b]{0.715\textwidth}
\centering
\includegraphics[width=\linewidth]{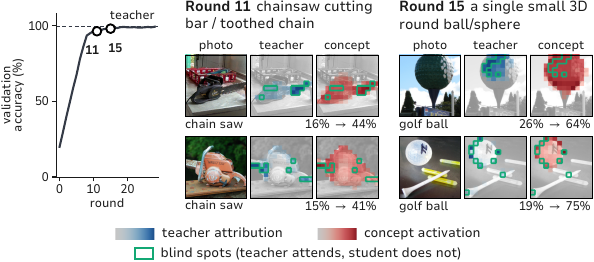}
\end{minipage}
\caption{\textit{Left:} \textbf{\method reaches near-full accuracy with a fraction of its concept budget.} Concepts needed to reach $90$/$95$/$98\%$ of the student's full-budget accuracy, per dataset; $K$ shown for reference. \textit{Right:} \textbf{Early rounds close large blind spots; later rounds refine with diminishing returns.} Round-by-round validation accuracy on Imagenette, with two exemplary rounds detailed.}
\label{fig:cak}
\label{fig:rounds_imagenette}
\label{fig:cak}
\label{fig:rounds_imagenette}
\end{figure}

%% file: content/figures/adaptive_examples.tex
\begin{figure}[t]
    \centering
    \includegraphics[width=0.95\textwidth]{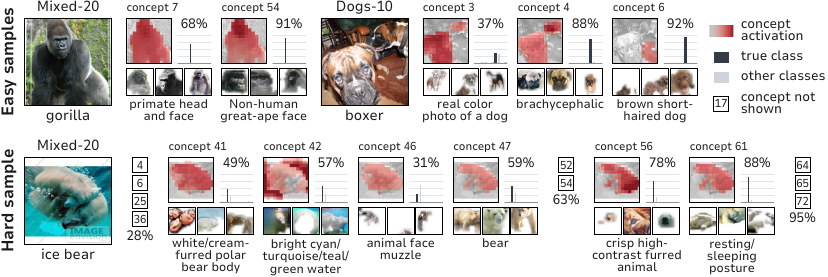}
    \caption{\textbf{Adaptive inference uses more concepts only on samples that need them.} Two easy samples (top) and one hard sample (bottom) given the task and selected concept set. For each sample, the activating concepts to reach 90\% confidence are shown, together with three exemplars. Boxed numbers mark firing concepts left out for visibility.}
    \label{fig:adaptive_mixed}
\end{figure}

%% file: content/tables/ablation_selection.tex
\begin{wraptable}{tr!}{0.38\textwidth}
    \vspace{-0pt}
    \centering
    \caption{\textbf{Every component of \method's selection facilitates compact, consistent, and accurate concept sets.} Ablation over selection strategy, averaged over vision datasets. LS: least-squares scoring (vs.\ random selection). Loc: localized candidate search. PF: prefiltering.}
    \label{tab:ablation-selection}
    \resizebox{0.38\textwidth}{!}{%
    \begin{tabular}{@{}ccc rrl@{}}
    \toprule
    LS & Loc & PF & K $\downarrow$ & C$^2_\text{patch}$ $\uparrow$ & $\Delta$ $\uparrow$ (\%) \\
    \midrule
    {\color{ForestGreen}\checkmark} & {\color{ForestGreen}\checkmark} & {\color{ForestGreen}\checkmark} & $145.0$  & $0.362$ & $-0.4$ \\
    {\color{ForestGreen}\checkmark} & {\color{ForestGreen}\checkmark} & --                               & $165.7$  & $0.345$ & $-0.9$ \\
    {\color{ForestGreen}\checkmark} & --                               & --                               & $190.5$  & $0.314$ & $-1.3$ \\
    --                               & --                               & --                               & $231.3$   & $0.272$ & $-2.1$ \\
    \bottomrule
    \end{tabular}
    }
    \vspace{-10pt}
\end{wraptable}

%% file: content/tables/text_cls.tex
\begin{wraptable}{hr!}{0.6\textwidth}
    \vspace{-10pt}
    \centering
    \caption{\textbf{\method generalizes across domains and teacher paradigms.} Evaluation of \method on two text datasets with a trained mlp-clf teacher and a zero-shot verbalizer teacher, verb-clf. We report concept count $K$, AutoInterp score (AIS), teacher gap ($\Delta$), and absolute balanced accuracy (Acc).}
    \label{tab:main_results_text}
    \resizebox{0.6\textwidth}{!}{%
    \begin{tabular}{@{}llrrrr@{}}
    \toprule
     & Method & K $\downarrow$ & AIS $\uparrow$ & $\Delta$ (\%) $\uparrow$ & Acc (\%) $\uparrow$ \\
    \midrule
    \multirow{4}{*}{\rotatebox{90}{AG News}}
      & SAE probe          & $31477$ & $0.327{\scriptstyle\pm0.009}$ & $-0.12{\scriptstyle\pm0.38}$ & $90.42{\scriptstyle\pm0.38}$ \\
      & \method (mlp)      & $93.7{\scriptstyle\pm6.4}$ & $0.469{\scriptstyle\pm0.006}$ & $-0.18{\scriptstyle\pm0.25}$ & $90.36{\scriptstyle\pm0.25}$ \\
    \noalign{\vskip 3pt}\cdashline{2-6}[2pt/2pt]\noalign{\vskip 3pt}
      & SAE probe          & $9710$ & $0.157{\scriptstyle\pm0.001}$ & $+18.57{\scriptstyle\pm0.98}$ & $88.25{\scriptstyle\pm0.98}$ \\
      & \method (verb)     & $30.0{\scriptstyle\pm5.0}$ & $0.094{\scriptstyle\pm0.020}$ & $+0.82{\scriptstyle\pm1.89}$ & $70.50{\scriptstyle\pm1.89}$ \\
    \midrule
    \multirow{4}{*}{\rotatebox{90}{PubMed}}
      & SAE probe          & $25893$ & $0.453{\scriptstyle\pm0.000}$ & $-2.47{\scriptstyle\pm0.87}$ & $73.58{\scriptstyle\pm0.87}$ \\
      & \method (mlp)      & $100.0{\scriptstyle\pm2.6}$ & $0.526{\scriptstyle\pm0.004}$ & $-0.02{\scriptstyle\pm0.23}$ & $76.03{\scriptstyle\pm0.23}$ \\
    \noalign{\vskip 3pt}\cdashline{2-6}[2pt/2pt]\noalign{\vskip 3pt}
      & SAE probe          & $7853$ & $0.087{\scriptstyle\pm0.000}$ & $+43.04{\scriptstyle\pm1.66}$ & $70.92{\scriptstyle\pm1.66}$ \\
      & \method (verb)     & $23.3{\scriptstyle\pm2.5}$ & $0.146{\scriptstyle\pm0.012}$ & $+15.93{\scriptstyle\pm2.04}$ & $43.81{\scriptstyle\pm2.04}$ \\
    \bottomrule
    \end{tabular}%
    }
    \vspace{-15pt}
\end{wraptable}

%% file: content/sections/5_discussion.tex
\section{Discussion \& Limitations}\label{sec:discussion}
\method combines several known components into one recipe: SAE-based concept discovery, teacher-student distillation via OMP-style residual selection, nested-dropout prediction, and agentic, automated concept naming. Together, they yield task-specific CBMs that are compact and accurate without concept annotations. We discuss the main limitations below. 

\textbf{Faithfulness, Not Correctness.} A concept is selected for reducing disagreement with the teacher, not for being correct: a teacher relying on a shortcut is matched by a bottleneck that reproduces it. However, \method's small concept budget should make such shortcuts easier to spot and revise in a human-in-the-loop setting \citep{SchramowskiSTBH20, StammerSK21}.

\textbf{Limits of the Concept Pool, Backbone and Teacher.} \method's performance is bound by its backbone and pretrained components: it cannot find a concept absent from the SAE dictionary, even if the teacher relies on other dimensions. Task-aware fine-tuning helps thereby, but is harder with scarce target-domain data. And while \method can outperform a poor teacher to some extent, its performance remains tied to the teacher's guidance.
Because concepts are scalar, non-negative activations rather than embeddings, concept leakage is bounded, and discrete activations can reduce it further. Sophisticated automated naming, e.g., MAIA, arguably identifies concept meaning more accurately than coarse image-text similarity, which helps users spot unintended semantics. Its counterfactual boundary testing can further clarify what a discovered concept represents.
A separate concern is whether a concept correlating strongly with the label is a shortcut for a finer-grained one. We argue that this is not automatic: if a task's true separating factor is coarse, a coarse concept is a sufficient decomposition, not a shortcut. Since selection ties concept count to task difficulty by construction, it stops once the teacher's residual is explained.

\textbf{Computational Efficiency.} \method's iterative loop trains a predictor $f_\mathcal{S}$ each round, but training a linear head is cheap, so cost scales with rounds, not model complexity. Rounds can be reduced by adding multiple concepts at once or jump-starting selection for harder problems. This cost is offset downstream, where the smaller bottleneck substantially cuts the cost of concept naming.

%% file: content/sections/6_conclusion.tex
\section{Conclusion}\label{sec:conclusion}
We presented \method, a recipe for building compact, task-specific concept bottlenecks from only a frozen backbone and an unsupervised concept pool: teacher-guided selection yields compact concept sets, globally and per sample, and nested dropout adapts their size to what each prediction needs. Across diverse domains and datasets, \method matches or exceeds teacher accuracy with an order of magnitude fewer concepts than annotation-free baselines, using a single recipe that transfers across domains and modalities. 
Overall, \method shows that annotation-free concept bottlenecks need not trade compactness for coverage: a small, discovered vocabulary can be as expressive as a much larger, fixed one, while remaining tractable to inspect.

\textbf{Future work.} Several extensions follow from \autoref{sec:method}'s design choices. 
The concept pool need not be an SAE: any unsupervised decomposition exposing per-token activations, e.g.\ \citet{bagci2026interpretability}, is compatible. A logic-based head \citep{vemuri2026logiccbms} could replace the linear predictor, pairing naturally with compositional naming \citep{mucompositional2020} as an alternative to MAIA. The recipe also extends to other tasks as long as suitable teachers are available. 
\method could be extended with an explicit grounding-verification step, checking that a concept's evidence is not merely residual-reducing but genuinely correct, as \citet{fang2026finegrained} do for medical imaging.
Finally, human or automated verification of MAIA's naming remains open. 

\section*{Acknowledgments}
This work has benefited from the GRK 2853 ``Neuroexplicit Models of Language, Vision, and Action'' (Project No. 471607914).

The project was further supported by the ``ML2MT'' project from the Volkswagen Stiftung, by the German Research Foundation (DFG) under Germany’s Excellence Strategy (EXC 3066/1 ``The Adaptive Mind'', Project No. 533717223 and by the EU-funded ``TANGO'' project (EU Horizon 2023, GA No 57100431). 
It has benefited from the HMWK project Hessian.AI, and from the Cluster of Excellence ``Reasonable AI'' funded by the DFG under Germany’s Excellence Strategy EXC-3057.

%% file: content/sections/7_reproducibility.tex
\section*{Reproducibility statement}

Code and access to the concept visualiser will be made public soon.
Appendix~\ref{app:exp_setup} specifies datasets (\autoref{app:exp_data_setup}), backbones, teachers, SAE architectures, fine-tuning objectives, and hyperparameters for both modalities (\autoref{app:exp_model_setup}), and Appendix~\ref{app:exp_metric_setup} defines every reported metric precisely, including $\mathrm{C}@q$, $\mathrm{C}^2$, and the AutoInterp score. Appendix~\ref{app:exp_model_setup} also documents implementation details of every annotation-free CBM baseline and \ref{app:prompts} shows the prompts for MAIA evaluations. Appendix~\ref{sec:results_appendix} reports per-dataset breakdowns underlying every averaged result in the main text, the accuracy-gap stopping tolerance ablation (\autoref{app:delta-acc-ablation}), and the full text results (\autoref{app:text_detail}). All results are averaged over three seeds, with model selection throughout by held-out validation accuracy.

%% file: content/sections/8_ai_use_statement.tex
\section*{AI use statement}
We used generative AI tools for the following required-disclosure tasks: refining hypotheses and design or feedback on research methodology and experiments (e.g.\ framing and stress-testing the discussion points in Section~\ref{sec:discussion}, and aspects of ablation and evaluation design); implementing methods, via agentic coding that wrote and executed code for \method and the baselines; and interpreting results, assisting in reading experimental outputs when drafting the corresponding text. We did not use generative AI tools to generate synthetic datasets, help develop theoretical models or conceptual frameworks, formulate mathematical claims, or provide ingredients for or assist in writing proofs, and did not use generative AI tools for translation, dataset cleaning or reformatting, or qualitative/thematic data analysis.
Additionally, we used generative AI tools for the following recommended-disclosure tasks: suggesting experimental parameters; creating and modifying scientific figures; editing code; identifying relevant literature and formatting references; and drafting and restructuring parts of the manuscript (text, section structure, and tables) to improve readability.
The authors reviewed AI-generated text, figures, and a portion of the AI-generated and AI-executed code visually for plausibility; we did not systematically test or verify all AI-generated code, nor exhaustively review all of it. We take responsibility for the final content of this work, including text, claims, and artifacts produced with the aid of generative AI.

%% file: content/sections/9_supplementary.tex
\onecolumn
\begin{center}
\textbf{\large Supplementary Materials}
\end{center}
\setcounter{section}{0}
\renewcommand{\thesection}{\Alph{section}}

\section{Impact Statement}
\label{app:impact_statement}

This work aims to improve the interpretability of machine learning systems by producing concept bottleneck models that use fewer, more coherent concepts than existing annotation free approaches. Improved interpretability tools of this kind can support model debugging, regulatory compliance, and human oversight in deployed systems, including in specialized domains such as medical imaging, where our ISIC-7 evaluation is one example.

We note several considerations specific to this method. First, \method's concept bottleneck is trained to match a black box teacher's behavior rather than to independently verify correctness; as discussed in \autoref{sec:discussion}, if the teacher relies on a spurious correlation or shortcut, a faithful bottleneck may reproduce rather than expose that shortcut. Practitioners should not treat a compact, coherent bottleneck as proof that the underlying teacher is free of such issues. Second, concept names are generated by an automated agent without a mandatory human verification step in our experiments; we recommend human review before any such bottleneck is used to inform consequential decisions, particularly in safety critical domains. Third, as with any interpretability method, there is a risk that a plausible sounding concept based explanation is used to build unwarranted trust in a system's predictions, independent of whether the explanation is accurate; we see this as a general risk for the concept bottleneck literature rather than one specific to this work.

Our experiments use publicly available benchmark datasets (Imagenette, ImageNet derived subsets, Flowers-102, ISIC-7, AG News, PubMed-20k) and do not involve new data collection or human subjects. We are not aware of dual use concerns beyond those generally applicable to interpretability and machine learning research.



\section{Additional Related Works}
\label{sec:related_app}

\textbf{Automatic Interpretability Agents.} Assigning human-readable semantics to a grounded concept \citep{knab2026s} is typically done via a single, cheap pass, matching activating inputs against a fixed or open vocabulary \citep{baudissection2017, oikarinendissect2022} or a trained captioning model \citep{hernandezmilan2022}. MAIA \citep{shaham2025maia} instead replaces this with an agent that iteratively designs and runs experiments on a concept before committing to a description, at a substantially higher per-concept cost; comparable agentic prescreening has since been explored elsewhere \citep{kim2025because}. \method's compactness and its naming choice are not independent decisions: MAIA is impractical over an SAE dictionary of thousands of atoms, but affordable once only a handful of concepts survive selection, a connection we are not aware of prior CBM work making explicit, though \citep{rao2024discover, wittenmayer2026cfm, ding2025conceptsae} face the same tradeoff implicitly, choosing cheaper, non-agentic naming over their respective concept pools.

\section{Algorithm}
\label{sec:detailed_alg}

We provide an algorithmic overview of \method in \autoref{alg:anybottle}.

\begin{algorithm}[h]
\caption{\methodfull{}}
\label{alg:anybottle}
\begin{algorithmic}[1]
\State Cache teacher outputs $Y_T$ and concept activations $C$ over full concept pool \Comment{Section~\autoref{sec:method_pool}}
\State $R \gets Y_T$ mean-centered; $\mathcal{S} \gets \{\arg\max_{j \in \{1,\dots,M\}} \, \mathrm{score}(\mathbf{c}_j, R)\}$ \Comment{initial concept, scored against the intercept-only residual}
\Repeat \Comment{Section~\autoref{sec:method_selection}}
    \State $R \gets$ residual of least squares fit of $\mathcal{S}$ against $Y_T$
    \State $\mathcal{I} \gets$ inputs with largest residual $\|R_i\|_2$
    \State $\mathcal{J} \gets$ candidate concepts at teacher/student attribution \textit{blind-spots} on $\mathcal{I}$
    \State $j^\star \gets \arg\max_{j \in \mathcal{J}} \mathrm{score}(\mathbf{c}_j, R)$
    \State $\mathcal{S} \gets \mathcal{S} \cup \{j^\star\}$
    \State retrain current predictor $f_\mathcal{S}$ on $\tilde{C}_\mathcal{S}$; track best validation accuracy so far
\Until{no valid candidate, accuracy gap closed, accuracy/EV plateau reached, or round limit reached}
\State $\mathcal{S}^\star \gets$ concept set of the best validation-accuracy round 
\State $f \gets$ retrain from $f_{\mathcal{S}^\star}$ on $\tilde{C}_{\mathcal{S}^\star}$ with nested dropout \Comment{Section~\autoref{sec:method_nested}}
\State automatically name each concept in $\mathcal{S}^\star$, subject to human review \Comment{Section~\autoref{sec:method_naming}}
\State \Return $\mathcal{S}^\star, f$
\end{algorithmic}
\end{algorithm}

\section{Additional Results}
\label{sec:results_appendix}
In this section we provide results from additional evaluations.

\begin{figure}[t]
    \centering
    \includegraphics[width=\textwidth]{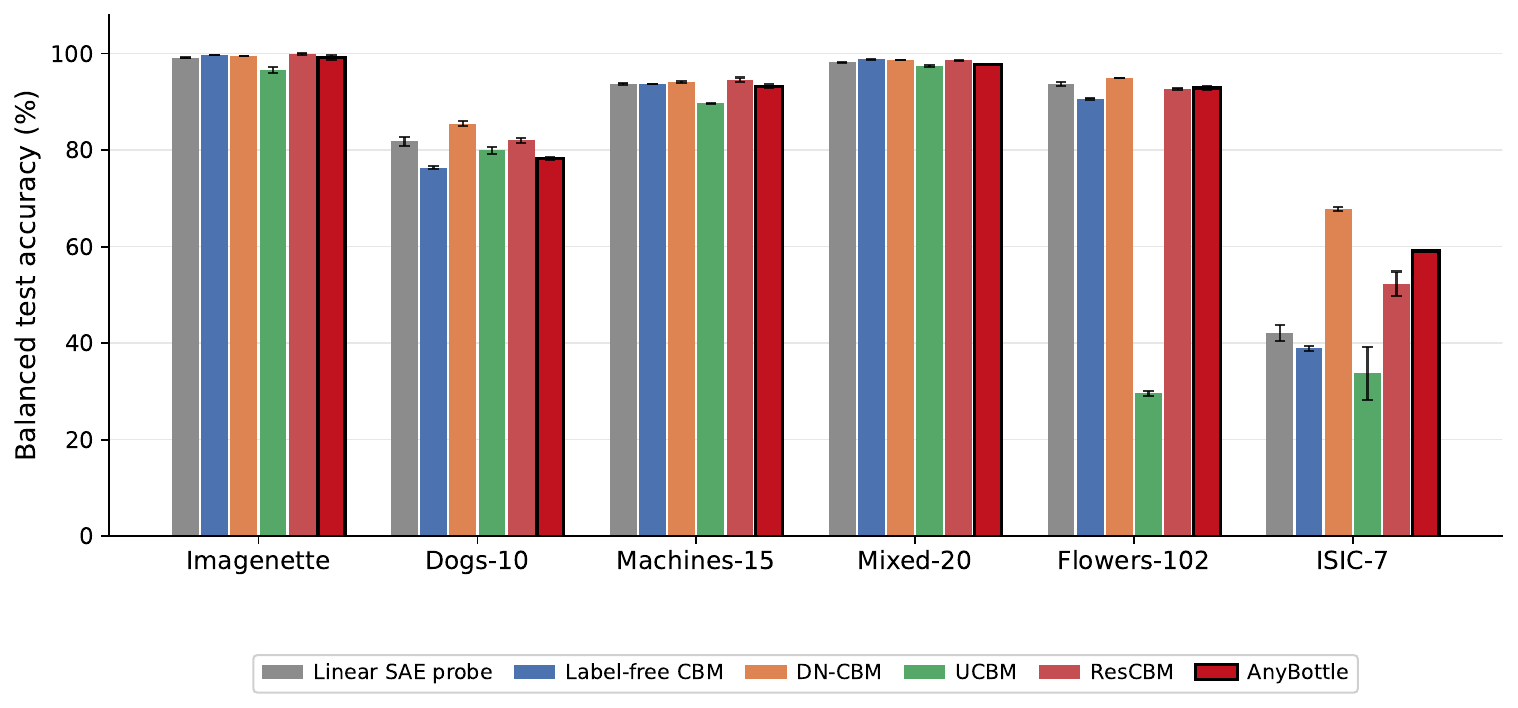}
    \caption{Raw accuracy of CBM baselines across vision datasets.}
    \label{fig:main_clf_acc}
\end{figure}

\subsection{Raw accuracy across vision datasets} \autoref{fig:main_clf_acc} reports raw balanced test accuracy for every method on every dataset, as a supplement to the teacher gap $\Delta$ reported in the main text (\autoref{fig:main}). Because each method uses its own backbone, these numbers are not directly comparable across methods: a higher bar does not imply a better bottleneck, since it may instead reflect a stronger or more task-suited backbone. We include this figure only to show that no method's raw accuracy is degenerate in absolute terms; the teacher gap $\Delta$, which normalizes for each method's own backbone, remains the primary accuracy metric throughout the paper. Accuracies are broadly comparable across methods on most datasets, with the exception of ISIC-7, where the spread across methods is largest.

\begin{figure}[t]
    \centering
    \includegraphics[width=\textwidth]{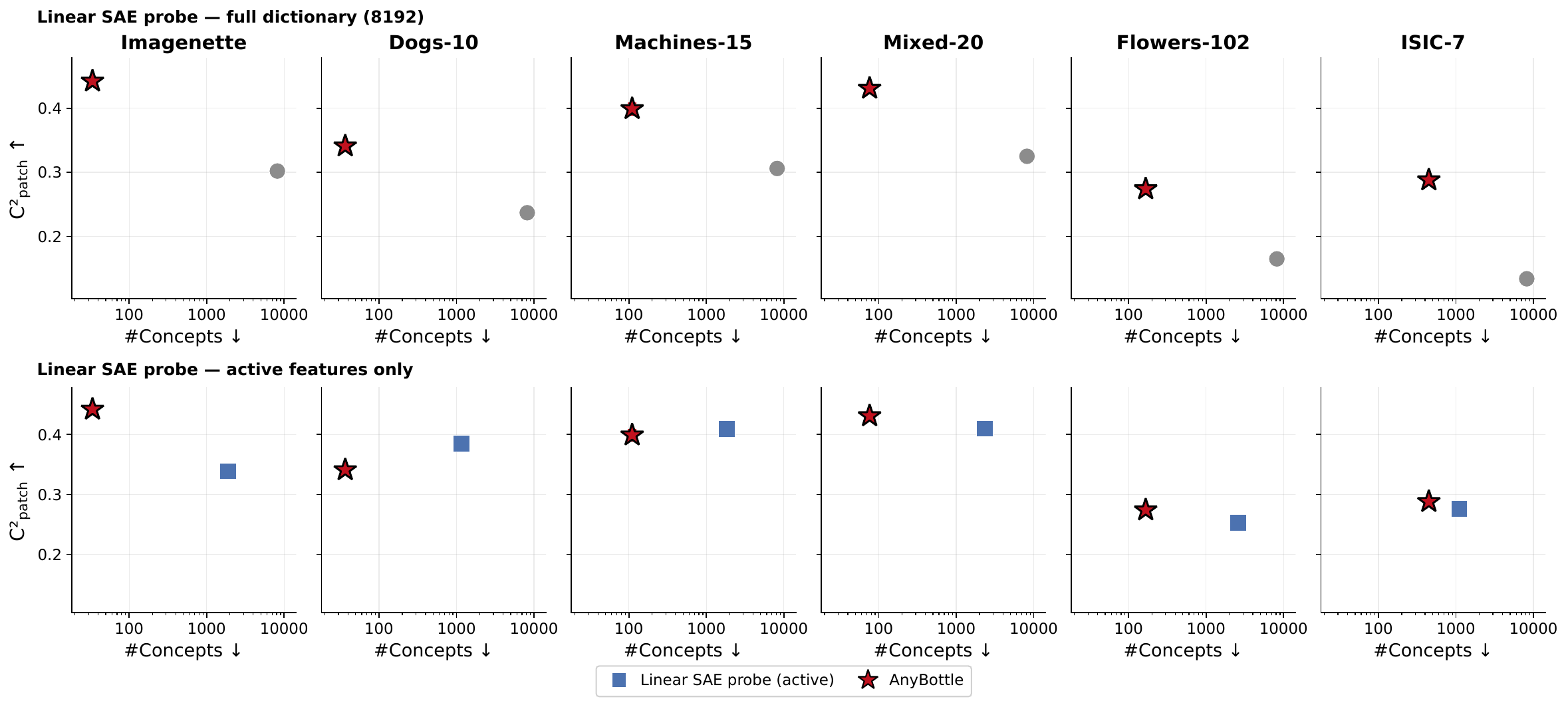}
    \caption{Patch-level concept consistency (C$^2_\text{patch}$) versus number of concepts, across six datasets. \emph{Top row}: Linear SAE probe evaluated at its full dictionary size (8192) compared to AnyBottle. \emph{Bottom row}: the same comparison restricted to the Linear SAE probe's active features only, shown at a finer x-axis scale. AnyBottle achieves comparable or higher C$^2_\text{patch}$ at substantially fewer concepts.}
    \label{fig:c2patch}
\end{figure}

\subsection{Patch-level concept consistency versus the linear SAE probe} \autoref{fig:c2patch} extends the image-level $\mathrm{C}^2$ comparison in the main text (\autoref{fig:main_results}) to the patch (region) level. This comparison is only possible against the linear SAE probe, not against the four CBM baselines: \method{} and the linear probe both operate on the same underlying SAE, whose concepts are grounded in individual patches, so patch-level activating evidence exists for every concept in the dictionary regardless of whether it was selected. The four CBM baselines produce only a single scalar activation per image per concept, with no sub-image localization, so no patch-level consistency score can be computed for them.

Top row: against the full, unselected dictionary (8192 features), \method{} achieves higher $\mathrm{C}^2_\text{patch}$ than the linear probe on every dataset, while using two to three orders of magnitude fewer concepts. Bottom row: restricting the linear probe to only its active features (\autoref{app:exp_metric_setup}) narrows the concept-count gap substantially, but \method{} still matches or exceeds the active-feature probe's consistency on four of six datasets (Imagenette, Mixed-20, Flowers-102, ISIC-7); Dogs-10 is the one dataset where the active-feature probe is clearly more consistent, and Machines-15 shows the two within each other's error bars. This indicates that \method's compactness is not simply discarding low-consistency concepts that an unconstrained active-feature baseline would also avoid, selection recovers comparable or better concept quality at a small fraction of the concept count on most, though not all, datasets.

\input{content/tables/ablation_selection_long}

\subsection{Per-Dataset Selection Strategy Ablation}
\autoref{tab:ablation-selection-per-dataset} reports the per-dataset values underlying the averaged ablation in \autoref{tab:ablation-selection}, showing concept set size $K$, C$^2_\textbf{patch}$ consistency and teacher accuracy gap $\Delta$. We ablate the different selection components: prefiltering (PF), localization (Loc), and candidate least squares scoring (LS).
C$^2_\text{patch}$ ordering is largely consistent across datasets: the full method has the highest consistency, and removing components generally reduces consistency. Concept count $K$ generally follows the same ordering: removing successive parts generally reduces compactness of the final concept selection, as less suited concepts are selected. On Mixed-20 and Dogs-10 the no-localization configuration selects a concept pool close in size to the full method's, despite the drop in consistency and in accuracy gap.
When only considering the teacher gap $\Delta$, there is no configuration that always outperforms over all datasets. While the full selection performs best on average, other configurations can perform slightly better on individual datasets.
Overall, the detailed per-dataset analysis confirms the findings from \autoref{tab:ablation-selection}. While the full selection mechanism is not unanimously better in all metrics for all datasets, removing individual components lowers the quality of the resulting concept selection in one or more aspects.

\input{content/tables/ablations_discrete_sae}

\subsection{Continuous and Discrete Bottlenecks}
\textbf{\method Supports Both Continuous and Discrete Bottlenecks.}
\autoref{tab:representation_taskaware} (left) compares discrete (binary) and continuous (percentile-normalised) concept representations. Both are natively supported by \method's selection procedure, and the choice can be made per deployment: discrete bottlenecks give the strictly present/absent semantics that much of the CBM literature builds on, useful for symbolic integration or settings where a binary interface is preferred, while continuous bottlenecks use fewer concepts on every dataset and match or improve both the teacher-accuracy gap and patch-level consistency on most of them. We use continuous representations by default, but the discrete variant remains a valid option within the same framework.

\subsection{Task-Aware SAE Fine-Tuning}
\textbf{Task-Aware SAE Fine-Tuning Helps Most Where the Pretrained Dictionary Is Mismatched.}
\autoref{tab:representation_taskaware} (right) compares the CC12M-pretrained SAE against the same SAE fine-tuned on-task via a KL loss to the teacher. On Imagenette, Machines-15, and Flowers-102, fine-tuning reduces concept count and improves consistency together; on Dogs-10 and Mixed-20, where the class differences are coarse and pretrained dictionary is already well matched, it offers less clear benefit. ISIC-7 is the clearest case: fine-tuning substantially closes the accuracy gap and improves consistency, at the cost of a larger concept budget, consistent with the pretrained dictionary being poorly matched to dermoscopic imagery. Across all six datasets, fine-tuning does not come at a systematic cost to concept consistency, indicating it adapts which features are selected. 

\input{content/tables/ablations_delta}

\subsection{Influence of the Accuracy-Gap Stopping Tolerance}
\label{app:delta-acc-ablation}

\autoref{tab:all-datasets-delta-acc-ablation} reports the effect of the stopping tolerance $\delta_{acc}$, which controls how much validation accuracy gap to the teacher is tolerated before selection halts and concepts are fixed. Tightening $\delta_{acc}$ from $0.0$ toward $0.2$ generally reduces the size of the selected concept sets, at a dataset-dependent cost to $\Delta_\text{val}$. On the other hand, in several scenarios changing the tolerance has no effect, since selection already halts on a different criterion first. Because model selection is driven by $\Delta_\text{val}$ rather than $\Delta_\text{test}$, a gap between the two is diagnostic of the stopping criterion fitting validation-specific noise rather than true generalization: on ISIC-7, attempting to match the validation performance of the teacher comes at a cost in test performance instead. This is most likely due to ISIC-7's weak, noisy teacher and small, unbalanced validation split (\autoref{sec:eval}). This suggests $\delta_{acc}$ is better treated as a budget-accuracy trade-off knob than as a target to tune aggressively: on well-behaved datasets it trades accuracy for compactness predictably, but where the teacher is unreliable, the validation signal can turn misleading exactly as the tolerance tightens.

\input{content/tables/text_cls_detailed}

\subsection{Generalization to Text in Detail}
\label{app:text_detail}

\autoref{tab:main_results_text_long} extends \autoref{tab:main_results_text} (main text) with $\mathrm{C}@90$ and $\mathrm{C}@95$, showing how much of each text dataset's concept budget $K$ is needed to reach near-full accuracy. The pattern differs by teacher paradigm. For the mlp-clf teacher, \method con reach most of its maximum accuracy with a smaller fraction of the concepts. For the verb-clf teacher, in contrast, nearly the entire (much smaller) concept pool is needed throughout on both datasets.

\FloatBarrier

\input{content/figures/rounds_all_datasets}
\input{content/figures/adaptive_supp}
\input{content/figures/concept_consistency}

\section{Qualitative Examples}
\label{sec:detaioed_qualitative}

This section extends the qualitative results in the main text (\autoref{sec:eval}) to all six vision datasets, and adds a further qualitative view of concept consistency not otherwise shown.

\textbf{Selection round by round, all datasets.} \autoref{fig:rounds_all} extends \autoref{fig:rounds_imagenette} (main text, Imagenette only) to all six datasets, showing the validation accuracy trajectory and three representative rounds' concepts and blind-spots per dataset. Note: Machines-15, Mixed-20, Flowers-102, and ISIC-7 use the class-focused jump-start round described in \autoref{sec:method_selection}, so their earliest round reflects this heuristic seed. Round numbers here are counted from the start of residual-driven selection that follows. The pattern from the main text holds throughout: early rounds close large, class-defining blind-spots, while later rounds refine coverage with diminishing accuracy gain. ISIC-7 stands out for needing far more rounds (314) than any other dataset to plateau, consistent with the results discussed in \autoref{sec:eval}.

\textbf{Adaptive inference, all datasets.} \autoref{fig:adaptive_supp} and \autoref{fig:adaptive_supp_2} extend \autoref{fig:adaptive_mixed} to all datasets, following the same selection and layout: for each dataset we show two correctly classified test images, one which needs a large number of firing concepts to reach 90\% confidence and which needs only a few. The same pattern recurs across datasets: easy images converge within a handful of concepts, while harder images draw on a much larger share of the discovered budget.

\textbf{Concept consistency, qualitatively.} \autoref{fig:consistency} and \autoref{fig:consistency_2} complement the quantitative $\mathrm{C}^2$ score with a direct visual check: five randomly drawn concepts per dataset, each shown with its ten most strongly activating test images. Most concepts are immediately recognizable and hold up across all ten images, not just the first one or two, e.g.\ Dogs-10's \emph{"erect pointed animal ear at top of head"} and Flowers-102's \emph{"round funnel/trumpet-shaped flower corolla"} pick out a consistent visual pattern throughout. A few concepts are visibly looser, most notably on ISIC-7, where patterns like \emph{"thin dark body hairs on pale skin"} and \emph{"pale normal skin with fine light micro-texture"} vary considerably in shape and framing across their top-activating images, and on Machines-15, where the broader \emph{"vehicles / means of transportation"} concept spans several vehicle types rather than one specific pattern. This mirrors the quantitative results, where ISIC-7 shows the lowest $\mathrm{C}^2$ scores of the six datasets (\autoref{fig:main_results}), consistent with its more specialized, harder-to-disentangle domain and the less adapted SAE concept pool.

These figures also highlight potential limitations of the used automated naming rather than of the concepts themselves: on Imagenette, the concept named \emph{"gas-station fuel nozzle / car fuel-filler inlet"} fires on several images whose gripped, handle-like protrusions more plausibly belong to a chainsaw's trigger housing than a fuel nozzle, a visual confusion likely inherited from Imagenette's own chain saw class. The concept's activating region is consistent, a handled mechanical fixture, but MAIA's single-pass name is more specific than the pattern it describes actually supports. We discuss such naming-accuracy risk in \autoref{sec:discussion}: compactness makes such cases tractable to catch on inspection, as here, but does not by itself guarantee the automatically generated name is correct, underscoring the value of human review before deployment.

\section{Details on Experimental Setup}
\label{app:exp_setup}

\subsection{Datasets}\label{app:exp_data_setup}

\paragraph{Vision.} We evaluate on six image classification datasets chosen to span a range of task types: \emph{Imagenette}, a coarse-grained 10-class subset of ImageNet; \emph{Dogs-10}, a fine-grained 10-breed subset requiring within-category discrimination; \emph{Machines-15}, a coarse multi-class task over mechanical object categories; \emph{Mixed-20}, a heterogeneous 20-class task combining categories from multiple sources; \emph{Flowers-102}, a large-vocabulary, fine-grained 102-class botanical task; and \emph{ISIC-7}, a 7-class dermoscopic skin lesion classification task representing a specialised scientific domain with a comparatively weak teacher. This selection deliberately covers coarse and fine-grained discrimination, small and large label vocabularies, and both generic and domain-specific imagery, to test whether concept selection behaves consistently across task difficulty rather than only on standard benchmarks.

\input{content/tables/imagenet_subsets}

\paragraph{ImageNet subsets.} To study how the difficulty of the discrimination task affects the learned bottleneck, we curate three subsets of ImageNet-1k~\citep{deng2009imagenet}. \textbf{ImageNet-Dogs} (10 classes) is a set of dog breeds. It includes groups of visually similar breeds (three retrievers, plus the Siberian husky and Eskimo dog) so that the task cannot be solved from coarse cues alone. \textbf{ImageNet-Machines} (15 classes) contains vehicles and machines from road, rail, water, air and industrial settings; while broader in scope than the dog breeds, it still groups several visually similar classes (e.g.\ sports car, convertible, and pickup, or the various trucks), giving it comparable fine-grained discrimination in parts of its label set. \textbf{ImageNet-Mixed} (20 classes) is a control condition that combines ten distinct animals with ten distinct machines. Table~\ref{tab:imagenet_subsets} lists the classes. For each subset we sample 700 training images per class and split them 90/10 into train and validation using a fixed seed (42). The official ImageNet validation images of the selected classes serve as the held-out test set.

\paragraph{Text.} We evaluate on two text classification datasets with different tasks and domains: \emph{AG News} (4-class topic classification, general-domain) and \emph{PubMed-20k} (5-class sentence role classification in biomedical abstracts, specialised domain). As in the vision setting, this spans a general-domain and a specialised, domain-specific text classification task.

\subsection{Model and Training Details}
\label{app:exp_model_setup}

\subsubsection{Vision Backbones and Teachers}
\method's vision backbone is CLIP-DINOiser~\citep{wysoczanska2024clip}: a CLIP ViT-B/16 model (\texttt{laion2b\_s34b\_b88k} weights) combined with DINO-style correlation pooling. This is a different backbone from the plain CLIP ViT-B/16 used by Label-free CBM and DN-CBM. It shares its CLIP weights (\texttt{laion2b}) with Res-CBM's substituted backbone, but the two backbones are otherwise different.

For every baseline and for \method, the teacher is a two-layer MLP (Linear$(d{\to}512)$--ReLU--Dropout$(0.1)$--Linear$(512{\to}C)$), trained on that method's own frozen backbone features via Adam (lr $10^{-3}$, weight decay $10^{-4}$, batch 256, 40 epochs), with model selection by balanced accuracy on a held-out validation split. Each baseline's teacher is trained separately, on that baseline's own backbone, so reported teacher--student gaps stay within one embedding space.

\paragraph{Label-free CBM~\citep{oikarinen2023label}.} Backbone: CLIP ViT-B/16 (\texttt{openai} weights), used both as the feature encoder and as the concept-scoring model, following the paper's \texttt{clip\_ViT-B/16} configuration. Concepts are generated once with GPT-4o-mini (the paper's original \texttt{text-davinci-002} is deprecated) using the paper's three prompt templates, then passed through the paper's five filtering steps (length, cosine similarity to the class name, cosine similarity to other concepts, mean CLIP activation, and post-projection similarity) unchanged. A linear projection layer is fit to maximize the CLIP-Dissect $\cos^3$ similarity objective, and the final sparse layer is fit with GLM-SAGA elastic-net regression ($\alpha=0.99$), tuned toward the paper's target of 25--35 non-zero weights per class.

\paragraph{Res-CBM~\citep{shang2024incremental}.} Backbone: CLIP ViT-B/16 (\texttt{laion2b} weights), chosen to keep this baseline in the same CLIP family as the rest of our comparisons rather than the original paper's RN50/ViT-L-14, so absolute numbers are not directly comparable to the published results. The candidate concept bank is built from Visual Genome object, attribute, and relationship terms, intersected with a CAV concept library and ConceptNet neighbours of the class names, following the paper's construction. Concepts are added incrementally over $D$ discovery rounds ($D$ set per dataset, following the paper's $\min(2C, 64)$ heuristic), each round initializing a new concept vector, training it, and mapping it to its nearest candidate via the paper's rank-based similarity loss.

\paragraph{DN-CBM~\citep{rao2024discover}.} Backbone: CLIP ViT-B/16 (\texttt{openai} weights), matching Label-free CBM's backbone so the two CLIP-based baselines are directly comparable to each other. Concepts come from the authors' released sparse autoencoder checkpoint (512$\to$4096, expansion factor 8, trained on CC3M), used frozen and forward-only, with concept names taken from the authors' released CLIP-Dissect-style name assignments. A single linear probe is trained on the resulting 4096-dimensional sparse code with an $\ell_1$ penalty, following the paper's ViT-B/16 configuration; because the SAE dictionary is large relative to the test sets, explained variance for this baseline is close to 1.0 and should be read as the concept space spanning the teacher's output space rather than as evidence of a tight fit.

\paragraph{UCBM~\citep{schrodi2024concept}.} Backbone: a frozen, ImageNet-supervised ResNet-50, used up to (excluding) global average pooling, following the paper's primary configuration; on our four ImageNet-derived datasets this backbone is in-distribution, so the teacher is near ceiling and the resulting gap is correspondingly small, while Flowers-102 and ISIC-7 are genuinely out-of-distribution for it. Concepts are discovered once per dataset via non-negative matrix factorization over cropped, class-conditioned activations, with the number of concepts per class set proportionally to the number of classes, following the paper's default factor. The classifier applies a cosine-similarity projection onto the discovered dictionary, an input-dependent gate that suppresses low-similarity concepts, concept dropout, and a sparse linear head, trained with the paper's elastic-net objective.

\paragraph{Comparability.} By default, each baseline uses a different backbone and concept source, and each has its own separately-trained teacher (above), so absolute accuracy is not comparable across baselines. For every baseline including \method, we report the teacher--student balanced accuracy gap and concept-count metrics within that baseline's own embedding space, under a common evaluation protocol: identical datasets, train/val/test splits, and three seeds (42, 123, 7) across all methods, with model selection by held-out validation balanced accuracy throughout (the original papers variously select on the training set, on a train/val split, or on the test set; we standardize all baselines, and \method, to a genuine held-out validation split for a symmetric comparison). We follow each paper's method code and hyperparameters as closely as possible.

\subsubsection{Vision SAE: Architecture and Fine-Tuning}
\method's concept pool is a Matryoshka BatchTopK SAE with activation dimension 512 and dictionary size 8192. It uses TopK activation with $k=12$ active latents per token, and has 6 nested Matryoshka groups. The SAE is trained on CLIP-DINOiser patch tokens over a CC12M corpus, with learning rate $10^{-4}$ for 118{,}935 steps. \method's selection procedure does not use the Matryoshka nesting structure and works with any SAE that exposes per-token activations over a fixed dictionary. The SAE's TopK activation ($k=12$) affects only its own reconstruction; it does not restrict \method's initial candidate pool, which always spans all 8192 dictionary atoms. 

We fine-tune one SAE checkpoint per dataset, in each case starting from the same CC12M-pretrained checkpoint. As detailed in \autoref{sec:method_pool}, we use the KL divergence of the teacher output distribution on standard embeddings and reconstructed SAE embeddings as objective for this fine-tuning.

\subsubsection{Text Backbones and Teachers}
We consider two teacher paradigms. The \emph{MLP teacher} (mlp-clf) mirrors the vision setup: a frozen, pretrained language model (Gemma-2-2B) is read via activation hooks, and a two-layer MLP head with one hidden layer of 512 units is trained on the model's last-token hidden state to predict the class label. Only this head is trained; the backbone stays frozen. The \emph{verbalizer teacher} (verb-clf) has no vision analogue: a frozen, instruction-tuned language model (Gemma-3-1B-it) is prompted directly, via a chat template, to name the class in natural language. Its prediction is read off the model's own next-token distribution, restricted to one single-token surface form per class (AG News: \emph{World}, \emph{Sports}, \emph{Business}, \emph{Technology}, the last standing in for the label \emph{Sci/Tech}; PubMed: \emph{Background}, \emph{Objective}, \emph{Methods}, \emph{Results}, \emph{Conclusion}) and renormalised over those tokens. Prompts are listed in \autoref{tab:verbalizer}.

\input{content/tables/verbalizer}

\subsubsection{Text SAE: Architecture per Teacher}
For the MLP-teacher (Gemma-2-2B), concepts come from a SAEBench-released TopK sparse autoencoder, dictionary size 64{,}512, read at layer 19. For the verbalizer-teacher (Gemma-3-1B-it), concepts come from a Gemma Scope sparse autoencoder, dictionary size 16{,}384, read at layer 17. This layer is fixed by configuration for all datasets. As on vision, we fine-tune one SAE checkpoint per dataset from a shared pretrained starting point, here using the \texttt{teacher\_kl} objective, again selected over the alternative fine-tuning modes by validation-set performance.
A redundancy filter discards candidate concepts whose SAE decoder direction is too cosine-similar (threshold $0.6$) to an already-selected concept due to the overly large SAE concept pool.

\subsubsection{Linear Probe and Full-Dictionary Baselines}
For both modalities, we compare against a linear probe trained on the full, unselected SAE dictionary, and one restricted to only the neurons active on the evaluation data.

\subsubsection{Hyperparameters}
\autoref{tab:hyperparams} lists the remaining hyperparameters governing selection and training, for vision and text where they differ.

\input{content/tables/hyperparams}

\subsubsection{Agentic Concept Naming}
\label{sec:maia}
We use the open-source OpenMAIA implementation \citep{camunas2025openmaia} with Qwen3-VL-8B-Instruct as both the interpretability agent and the image-description model, served locally via vLLM (32k context). For each selected concept, MAIA runs its full agentic loop: in every round it writes Python experiments that are executed against our system, with access to the concept's activations and per-patch activation heatmaps on arbitrary images, its 15 top-activating images from the held-out probe set, and captioning. It can also synthesize and edit images with FLUX.1-dev and FLUX.1-Kontext-dev (4-bit quantized; 25 and 15 denoising steps, guidance 5.5 and 2.5) to test hypotheses causally. The loop ends when the agent returns a final description; after 25 rounds it is prompted to conclude. The agent outputs a short label, a description, and a self-reported confidence score. We provide prompts in \autoref{app:prompts}.

\subsection{Metrics}
\label{app:exp_metric_setup}

Since each baseline uses a different backbone, concept source, and teacher, absolute accuracy is not comparable across methods. We instead report each method's own teacher student accuracy gap, explained variance, and concept count within its own embedding space, under an identical protocol: identical datasets, train/val/test splits, and three seeds across all methods.

\paragraph{Number of concepts.} We report $K = |\mathcal{S}^\star|$, the size of the concept set returned by selection, as the primary measure of bottleneck compactness. For full dictionary SAE comparisons (the linear SAE probe baseline), we report the number of active SAE features in place of $K$ (see below). Enabled by the nested dropout head, ?method's final head $f$ can predict at any prefix length $k \leq K$. After training, $f$ is evaluated across a range of prefix lengths on both validation and test data, producing a balanced accuracy versus concept budget profile $\mathrm{acc}(k)$, where $\mathrm{acc}(\cdot)$ denotes balanced accuracy throughout, and $\mathrm{acc}(K)$ is reported as the model's overall accuracy. We define
\begin{equation}
    \mathrm{C}@q = \min \left\{ k : \mathrm{acc}(k) \geq \frac{q}{100} \cdot \mathrm{acc}(K) \right\},
\end{equation}
the smallest prefix length at which test balanced accuracy first reaches $q$ percent of the student's own full-budget accuracy $\mathrm{acc}(K)$, for $q \in \{90, 95, 98\}$, characterising how much of the discovered budget is actually needed to reach near-ceiling performance, rather than how many concepts were discovered in total. Unlike C@90/95/98, $K$ is not itself an accuracy milestone: it is the total selected budget, reported alongside C@q for reference.

\paragraph{Active SAE features.} The full-dictionary linear SAE probe uses every SAE feature, so we report its number of \emph{active} features in place of $K$. 
A feature counts as active if it fires on at least 5 examples of the held-out test split. It fires on an example if its max-pooled activation is above zero. We use the same active subset as the probe's reference set for the $\mathrm{C}^2$ and AIS comparisons, and the AIS features are sampled from this subset, stratified by activation frequency. This subset is computed after the probe has been trained and is used only for reporting and evaluation: the probe is trained on, and evaluated with, the full dictionary. 

\paragraph{Explained variance (EV).} As a running diagnostic during selection, we track $\mathrm{EV}_S = 1 - \mathrm{var}(R)/\mathrm{var}(Y_T)$, the fraction of teacher output variance explained by the current concept set $S$. We report EV alongside $K$ and accuracy where relevant, as an indicator of concept pool quality independent of the stopping criterion reached.

\paragraph{Balanced accuracy.} As several datasets have substantial class imbalance, we report balanced accuracy, the average of per-class recall
\begin{equation}
    \mathrm{BAcc} = \frac{1}{|\mathcal{Y}|} \sum_{c \in \mathcal{Y}} \frac{\text{correctly predicted}_c}{\text{total}_c}.
\end{equation}

\paragraph{Concept consistency ($\mathrm{C}^2$).} Accuracy alone does not indicate whether the selected concepts are individually coherent, i.e. whether each concept's high-activation examples share a consistent, describable pattern rather than firing on an arbitrary mix of inputs. We adopt the $\mathrm{C}^2$ score~\citep{parchami2026fact} to quantify this: for each selected concept, $\mathrm{C}^2$ measures the semantic consistency of the set of inputs (or input regions) that most strongly activate it, and we report the mean $\mathrm{C}^2$ over the selected concepts. For \method, this is computed at the \emph{patch} level, since each SAE concept's activating evidence is localized to specific image patches. The four CBM baselines we compare against instead produce a single scalar activation per image per concept, either a text-concept cosine similarity or a global-feature projection, so their $\mathrm{C}^2$ is necessarily computed at the coarser \emph{image} level; we report this distinction alongside each result and note that region-level and image-level $\mathrm{C}^2$ are not directly comparable; the baseline scores serve to indicate whether a method's concepts are coherent \emph{at the granularity it operates at}, not to rank fine-grained localization ability the baselines were never designed to provide. 
We additionally compare \method against two references on its own SAE pool: a linear probe over the full, unselected SAE dictionary ($M$ concepts), and a linear probe restricted to only the subset of SAE neurons that are ever active on the evaluation data (\emph{active} neurons), both trained under the same protocol as the student. 
Together with $K$, $\mathrm{C}^2$ lets us distinguish genuine gains in concept quality from AnyBottle's selection procedure from gains that would follow from using any smaller, arbitrary subset of the dictionary.

\paragraph{AutoInterp score (AIS).} For text, where concepts do not localize to spatial regions, we assess concept coherence following the automated interpretability protocol of \citep{bills2023language,paulo2024automatically}. For each concept, an LLM judge is shown the concept's top activating documents from the evaluation split and produces a single sentence description of the pattern. The same judge is then given only this description, together with a held out set of ten strongly activating and ten weakly or non activating documents, and rates per document how strongly it expects the concept to fire. The AutoInterp score for a concept is the Pearson correlation between these predicted ratings and the concept's true activations on the same held out documents; a concept is excluded from the aggregate, rather than scored as zero, when this correlation is undefined (fewer than two held out documents, or zero variance in predicted or true activations). We report AIS as the unweighted mean over concepts with a defined value. The judge is Qwen2.5 14B Instruct, decoded greedily for determinism. For \method, AIS covers the discovered concept set $\mathcal{S}^\star$ in full. For the linear SAE probe baseline, scoring the full dictionary is infeasible. We instead restrict to the active subset (activation greater than zero on at least five test documents), partition it into five quantile bins by document activation frequency, and draw 200 latents without replacement, split evenly across bins. This stratified draw is repeated over 20 independent seeds, and we report the mean and standard deviation of AIS across the 20 draws in our provided results.

\subsection{Prompts}
\label{app:prompts}

\input{content/prompts/maia_prompts}

%% file: content/tables/ablation_selection_long.tex
\begin{table*}[t]
\centering
\caption{Per-dataset results underlying the averaged ablation in
Table~\autoref{tab:ablation-selection} (main text). Ablations of least-squares scoring (LS), localized
candidate search (Loc), and prefiltering (PF). $\Delta$ = balanced test accuracy $-$
balanced teacher accuracy. C$^2_\text{patch}$ is the patch-level concept
consistency. Mean$\pm$std over 3 seeds.}
\label{tab:ablation-selection-per-dataset}
\resizebox{0.7\textwidth}{!}{%
\begin{tabular}{@{}lccc rrl@{}}
\toprule
Dataset & LS & Loc & PF & K $\downarrow$ & C$^2_\text{patch}$ $\uparrow$ & $\Delta$ $\uparrow$ (\%) \\
\midrule
\multirow{4}{*}{Imagenette}
  & {\color{ForestGreen}\checkmark} & {\color{ForestGreen}\checkmark} & {\color{ForestGreen}\checkmark} & $34.0{\scriptstyle\pm2.0}$   & $0.442{\scriptstyle\pm0.005}$ & $+0.3{\scriptstyle\pm0.6}$ \\
  & {\color{ForestGreen}\checkmark} & {\color{ForestGreen}\checkmark} & --                               & $47.3{\scriptstyle\pm3.5}$   & $0.402{\scriptstyle\pm0.008}$ & $-0.9{\scriptstyle\pm0.4}$ \\
  & {\color{ForestGreen}\checkmark} & --                               & --                               & $87.3{\scriptstyle\pm1.5}$   & $0.348{\scriptstyle\pm0.003}$ & $+0.4$ \\
  & --                               & --                               & --                               & $184.7{\scriptstyle\pm22.8}$ & $0.269{\scriptstyle\pm0.015}$ & $+0.2$ \\
\midrule
\multirow{4}{*}{Dogs-10}
  & {\color{ForestGreen}\checkmark} & {\color{ForestGreen}\checkmark} & {\color{ForestGreen}\checkmark} & $37.0{\scriptstyle\pm1.0}$   & $0.341{\scriptstyle\pm0.002}$ & $-1.8{\scriptstyle\pm0.3}$ \\
  & {\color{ForestGreen}\checkmark} & {\color{ForestGreen}\checkmark} & --                               & $49.0{\scriptstyle\pm0.0}$   & $0.345{\scriptstyle\pm0.000}$ & $-2.0{\scriptstyle\pm0.2}$ \\
  & {\color{ForestGreen}\checkmark} & --                               & --                               & $38.3{\scriptstyle\pm1.2}$   & $0.331{\scriptstyle\pm0.000}$ & $-2.3$ \\
  & --                               & --                               & --                               & $128.7{\scriptstyle\pm20.6}$ & $0.280{\scriptstyle\pm0.011}$ & $-1.4$ \\
\midrule
\multirow{4}{*}{Machines-15}
  & {\color{ForestGreen}\checkmark} & {\color{ForestGreen}\checkmark} & {\color{ForestGreen}\checkmark} & $110.0{\scriptstyle\pm14.0}$ & $0.399{\scriptstyle\pm0.009}$ & $+0.3{\scriptstyle\pm0.4}$ \\
  & {\color{ForestGreen}\checkmark} & {\color{ForestGreen}\checkmark} & --                               & $219.0{\scriptstyle\pm80.9}$ & $0.377{\scriptstyle\pm0.021}$ & $+0.8{\scriptstyle\pm0.4}$ \\
  & {\color{ForestGreen}\checkmark} & --                               & --                               & $167.3{\scriptstyle\pm40.3}$ & $0.344{\scriptstyle\pm0.016}$ & $+0.5$ \\
  & --                               & --                               & --                               & $265.0{\scriptstyle\pm10.6}$ & $0.308{\scriptstyle\pm0.006}$ & $+0.4$ \\
\midrule
\multirow{4}{*}{Mixed-20}
  & {\color{ForestGreen}\checkmark} & {\color{ForestGreen}\checkmark} & {\color{ForestGreen}\checkmark} & $76.0{\scriptstyle\pm3.0}$   & $0.431{\scriptstyle\pm0.006}$ & $-0.3{\scriptstyle\pm0.1}$ \\
  & {\color{ForestGreen}\checkmark} & {\color{ForestGreen}\checkmark} & --                               & $137.7{\scriptstyle\pm2.1}$  & $0.423{\scriptstyle\pm0.000}$ & $+0.1{\scriptstyle\pm0.1}$ \\
  & {\color{ForestGreen}\checkmark} & --                               & --                               & $78.0{\scriptstyle\pm3.5}$   & $0.408{\scriptstyle\pm0.005}$ & $-0.6$ \\
  & --                               & --                               & --                               & $150.0{\scriptstyle\pm25.2}$ & $0.335{\scriptstyle\pm0.012}$ & $-0.9$ \\
\midrule
\multirow{4}{*}{Flowers-102}
  & {\color{ForestGreen}\checkmark} & {\color{ForestGreen}\checkmark} & {\color{ForestGreen}\checkmark} & $166.0{\scriptstyle\pm10.0}$ & $0.274{\scriptstyle\pm0.006}$ & $+1.1{\scriptstyle\pm0.4}$ \\
  & {\color{ForestGreen}\checkmark} & {\color{ForestGreen}\checkmark} & --                               & $156.0{\scriptstyle\pm9.2}$  & $0.270{\scriptstyle\pm0.003}$ & $+0.7{\scriptstyle\pm0.4}$ \\
  & {\color{ForestGreen}\checkmark} & --                               & --                               & $291.0{\scriptstyle\pm4.4}$  & $0.245{\scriptstyle\pm0.000}$ & $+0.4$ \\
  & --                               & --                               & --                               & $230.7{\scriptstyle\pm14.6}$ & $0.235{\scriptstyle\pm0.009}$ & $+1.0$ \\
\midrule
\multirow{4}{*}{ISIC-7}
  & {\color{ForestGreen}\checkmark} & {\color{ForestGreen}\checkmark} & {\color{ForestGreen}\checkmark} & $447.0{\scriptstyle\pm30.0}$ & $0.288{\scriptstyle\pm0.001}$ & $-2.2{\scriptstyle\pm0.2}$ \\
  & {\color{ForestGreen}\checkmark} & {\color{ForestGreen}\checkmark} & --                               & $385.3{\scriptstyle\pm67.8}$ & $0.254{\scriptstyle\pm0.000}$ & $-3.9{\scriptstyle\pm0.6}$ \\
  & {\color{ForestGreen}\checkmark} & --                               & --                               & $481.0{\scriptstyle\pm12.5}$ & $0.205{\scriptstyle\pm0.001}$ & $-6.0$ \\
  & --                               & --                               & --                               & $428.3{\scriptstyle\pm90.4}$ & $0.208{\scriptstyle\pm0.006}$ & $-11.7$ \\
\bottomrule
\end{tabular}%
}
\end{table*}

%% file: content/tables/ablations_discrete_sae.tex
\begin{table*}[t]
\centering
\caption{Ablation studies on concept representation (left) and SAE fine-tuning (right).
\textbf{Left}: effect of concept representation with a Task-aware: discrete vs continuous concept scores. \textbf{Right}: effect of SAE fine-tuning with
continuous concepts: task-agnostic pretrained vs. task-aware finetuned. 
$\Delta$ = balanced test accuracy $-$ balanced teacher accuracy;
C$^2_\text{patch}$ is the patch-level concept consistency. Mean$\pm$std over 3 seeds. }
\label{tab:representation_taskaware}
\begin{minipage}[t]{0.49\textwidth}
\centering
\resizebox{\textwidth}{!}{%

\begin{tabular}{@{}llrrl@{}}
\toprule
Dataset & Config & \#Concepts $\downarrow$ & C$^2_\text{patch}$ $\uparrow$ & $\Delta$ $\uparrow$ (\%) \\
\midrule
\multirow{3}{*}{Imagenette}
  & Discrete & $90.0{\scriptstyle\pm9.0}$ & $0.38{\scriptstyle\pm0.012}$ & $+0.1{\scriptstyle\pm0.4}$ \\
  & Continuous & $34.0{\scriptstyle\pm2.0}$ & $0.442{\scriptstyle\pm0.005}$ & $+0.3{\scriptstyle\pm0.6}$ \\
\midrule
\multirow{3}{*}{Dogs-10}
  & Discrete & $89.0{\scriptstyle\pm3.0}$ & $0.317{\scriptstyle\pm0.004}$ & $+0.3{\scriptstyle\pm0.8}$ \\
  & Continuous & $37.0{\scriptstyle\pm1.0}$ & $0.341{\scriptstyle\pm0.002}$ & $-1.8{\scriptstyle\pm0.3}$ \\
\midrule
\multirow{3}{*}{Machines-15}
  & Discrete & $141.0{\scriptstyle\pm38.0}$ & $0.405{\scriptstyle\pm0.015}$ & $-0.9{\scriptstyle\pm1.1}$ \\
  & Continuous & $110.0{\scriptstyle\pm14.0}$ & $0.399{\scriptstyle\pm0.009}$ & $+0.3{\scriptstyle\pm0.4}$ \\
\midrule
\multirow{3}{*}{Mixed-20}
  & Discrete & $79.0{\scriptstyle\pm5.0}$ & $0.443{\scriptstyle\pm0.006}$ & $-1.2{\scriptstyle\pm0.3}$ \\
  & Continuous & $76.0{\scriptstyle\pm3.0}$ & $0.431{\scriptstyle\pm0.006}$ & $-0.3{\scriptstyle\pm0.1}$ \\
\midrule
\multirow{3}{*}{Flowers-102}
  & Discrete & $335.0{\scriptstyle\pm24.0}$ & $0.25{\scriptstyle\pm0.001}$ & $+0.7{\scriptstyle\pm0.2}$ \\
  & Continuous & $166.0{\scriptstyle\pm10.0}$ & $0.274{\scriptstyle\pm0.006}$ & $+1.1{\scriptstyle\pm0.4}$ \\
\midrule
\multirow{3}{*}{ISIC-7}
  & Discrete & $465.0{\scriptstyle\pm55.0}$ & $0.285{\scriptstyle\pm0.003}$ & $-4.6{\scriptstyle\pm2.1}$ \\
  & Continuous & $447.0{\scriptstyle\pm30.0}$ & $0.288{\scriptstyle\pm0.001}$ & $-2.2{\scriptstyle\pm0.2}$ \\
\bottomrule
\end{tabular}%
}
\end{minipage}
\hfill
\begin{minipage}[t]{0.49\textwidth}
\centering
\resizebox{\textwidth}{!}{%

\begin{tabular}{@{}llrrl@{}}
\toprule
Dataset & Config & \#Concepts $\downarrow$ & C$^2_\text{patch}$ $\uparrow$ & $\Delta$ $\uparrow$ (\%) \\
\midrule
\multirow{2}{*}{Imagenette}
  & Pretrained & $81.0{\scriptstyle\pm1.0}$ & $0.306{\scriptstyle\pm0.0}$ & $-0.1{\scriptstyle\pm0.3}$ \\
  & Task-aware & $34.0{\scriptstyle\pm2.0}$ & $0.442{\scriptstyle\pm0.005}$ & $+0.3{\scriptstyle\pm0.6}$ \\
\midrule
\multirow{2}{*}{Dogs-10}
  & Pretrained & $36.0{\scriptstyle\pm9.0}$ & $0.388{\scriptstyle\pm0.029}$ & $-1.4{\scriptstyle\pm0.9}$ \\
  & Task-aware & $37.0{\scriptstyle\pm1.0}$ & $0.341{\scriptstyle\pm0.002}$ & $-1.8{\scriptstyle\pm0.3}$ \\
\midrule
\multirow{2}{*}{Machines-15}
  & Pretrained & $182.0{\scriptstyle\pm6.0}$ & $0.355{\scriptstyle\pm0.003}$ & $+0.4{\scriptstyle\pm0.5}$ \\
  & Task-aware & $110.0{\scriptstyle\pm14.0}$ & $0.399{\scriptstyle\pm0.009}$ & $+0.3{\scriptstyle\pm0.4}$ \\
\midrule
\multirow{2}{*}{Mixed-20}
  & Pretrained & $58.0{\scriptstyle\pm0.0}$ & $0.464{\scriptstyle\pm0.0}$ & $-0.5{\scriptstyle\pm0.2}$ \\
  & Task-aware & $76.0{\scriptstyle\pm3.0}$ & $0.431{\scriptstyle\pm0.006}$ & $-0.3{\scriptstyle\pm0.1}$ \\
\midrule
\multirow{2}{*}{Flowers-102}
  & Pretrained & $404.0{\scriptstyle\pm138.0}$ & $0.24{\scriptstyle\pm0.002}$ & $-1.9{\scriptstyle\pm0.6}$ \\
  & Task-aware & $166.0{\scriptstyle\pm10.0}$ & $0.274{\scriptstyle\pm0.006}$ & $+1.1{\scriptstyle\pm0.4}$ \\
\midrule
\multirow{2}{*}{ISIC-7}
  & Pretrained & $358.0{\scriptstyle\pm27.0}$ & $0.263{\scriptstyle\pm0.001}$ & $-12.4{\scriptstyle\pm0.4}$ \\
  & Task-aware & $447.0{\scriptstyle\pm30.0}$ & $0.288{\scriptstyle\pm0.001}$ & $-2.2{\scriptstyle\pm0.2}$ \\
\bottomrule
\end{tabular}%
}
\end{minipage}
\end{table*}

%% file: content/tables/ablations_delta.tex
\begin{table}[t]
\centering
\caption{Effect of the stopping tolerance $\delta_{acc}$ on AnyBottle, across all six vision datasets. $\Delta$ = Accuracy $-$ Teacher (balanced accuracy). C$^2_\text{patch}$ is the patch-level (localised) consistency score; C$^2_\text{img}$ is the image-level score. Mean$\pm$std over 3 seeds. }
\label{tab:all-datasets-delta-acc-ablation}
\resizebox{0.6\textwidth}{!}{%
\begin{tabular}{@{}llrrrrr@{}}
\toprule
Dataset & $\delta_{acc}$ & \#Concepts & C$^2_\text{patch}$ $\uparrow$ & C$^2_\text{img}$ $\uparrow$ & $\Delta_\text{val}$ (\%) & $\Delta_\text{test}$ (\%) \\
\midrule
\multirow{3}{*}{Imagenette }
  & $0.0$ & $33.7{\scriptstyle\pm2.1}$ & $0.442{\scriptstyle\pm0.005}$ & $0.324{\scriptstyle\pm0.003}$ & $+0.09$ & $+0.33$ \\
  & $0.1$ & $33.7{\scriptstyle\pm2.1}$ & $0.442{\scriptstyle\pm0.005}$ & $0.324{\scriptstyle\pm0.003}$ & $+0.09$ & $+0.33$ \\
  & $0.2$ & $25.0{\scriptstyle\pm4.6}$ & $0.453{\scriptstyle\pm0.007}$ & $0.335{\scriptstyle\pm0.007}$ & $-0.19$ & $-0.27$ \\
\midrule\midrule
\multirow{3}{*}{Dogs-10 }
  & $0.0$ & $36.7{\scriptstyle\pm0.6}$ & $0.341{\scriptstyle\pm0.002}$ & $0.105{\scriptstyle\pm0.001}$ & $+0.05$ & $-1.80$ \\
  & $0.1$ & $36.7{\scriptstyle\pm0.6}$ & $0.341{\scriptstyle\pm0.002}$ & $0.105{\scriptstyle\pm0.001}$ & $+0.05$ & $-1.80$ \\
  & $0.2$ & $36.7{\scriptstyle\pm0.6}$ & $0.341{\scriptstyle\pm0.002}$ & $0.105{\scriptstyle\pm0.001}$ & $+0.05$ & $-1.80$ \\
\midrule\midrule
\multirow{3}{*}{Machines-15 }
  & $0.0$ & $143.0{\scriptstyle\pm66.2}$ & $0.391{\scriptstyle\pm0.033}$ & $0.233{\scriptstyle\pm0.028}$ & $+0.10$ & $+0.27$ \\
  & $0.1$ & $110.3{\scriptstyle\pm13.7}$ & $0.399{\scriptstyle\pm0.009}$ & $0.234{\scriptstyle\pm0.009}$ & $+0.48$ & $+0.31$ \\
  & $0.2$ & $89.0{\scriptstyle\pm9.5}$ & $0.414{\scriptstyle\pm0.010}$ & $0.254{\scriptstyle\pm0.011}$ & $-0.06$ & $-0.57$ \\
\midrule\midrule
\multirow{3}{*}{Mixed-20 }
  & $0.0$ & $82.0{\scriptstyle\pm4.0}$ & $0.406{\scriptstyle\pm0.009}$ & $0.303{\scriptstyle\pm0.008}$ & $+0.19$ & $-0.60$ \\
  & $0.1$ & $76.3{\scriptstyle\pm3.1}$ & $0.431{\scriptstyle\pm0.006}$ & $0.325{\scriptstyle\pm0.006}$ & $+0.08$ & $-0.30$ \\
  & $0.2$ & $71.3{\scriptstyle\pm3.8}$ & $0.428{\scriptstyle\pm0.007}$ & $0.321{\scriptstyle\pm0.005}$ & $+0.00$ & $-0.63$ \\
\midrule\midrule
\multirow{3}{*}{Flowers-102 }
  & $0.0$ & $197.3{\scriptstyle\pm13.3}$ & $0.265{\scriptstyle\pm0.002}$ & $0.152{\scriptstyle\pm0.002}$ & $+1.37$ & $+1.59$ \\
  & $0.1$ & $166.3{\scriptstyle\pm10.1}$ & $0.274{\scriptstyle\pm0.006}$ & $0.158{\scriptstyle\pm0.003}$ & $+0.78$ & $+1.10$ \\
  & $0.2$ & $188.0{\scriptstyle\pm3.5}$ & $0.267{\scriptstyle\pm0.001}$ & $0.154{\scriptstyle\pm0.001}$ & $+0.91$ & $+1.38$ \\
\midrule\midrule
\multirow{3}{*}{ISIC-7 }
  & $0.0$ & $447.0{\scriptstyle\pm29.6}$ & $0.288{\scriptstyle\pm0.001}$ & $0.088{\scriptstyle\pm0.000}$ & $-3.08$ & $-2.17$ \\
  & $0.1$ & $447.0{\scriptstyle\pm29.6}$ & $0.288{\scriptstyle\pm0.001}$ & $0.088{\scriptstyle\pm0.000}$ & $-3.08$ & $-2.17$ \\
  & $0.2$ & $404.3{\scriptstyle\pm91.0}$ & $0.290{\scriptstyle\pm0.004}$ & $0.088{\scriptstyle\pm0.001}$ & $-8.75$ & $-2.53$ \\
\bottomrule
\end{tabular}%
}
\end{table}

%% file: content/tables/text_cls_detailed.tex
\begin{table*}[t]
\centering
\caption{Across domains and teacher paradigms, \method{} matches near-teacher accuracy with orders of magnitude fewer concepts than a linear probe over the full SAE dictionary. We report concept count $K$, concepts needed to reach $90\%$ and $95\%$ of the student accuracy (C@90/95), AutoInterp score (AIS), and balanced accuracy gap to teacher ($\Delta$) and in absolute terms (Acc).}
\label{tab:main_results_text_long}
\resizebox{\textwidth}{!}{%
\begin{tabular}{@{}llrrrrrr@{}}
\toprule
Dataset & Method & \#Concepts $\downarrow$ & C@90 $\downarrow$ & C@95 $\downarrow$ & AIS $\uparrow$ & $\Delta$ (\%) $\uparrow$ & Acc (\%) $\uparrow$ \\
\midrule
\multirow{4}{*}{AG News}
  & Linear SAE probe         & $31477$ & -- & -- & $0.327{\scriptstyle\pm0.009}$ & $-0.12{\scriptstyle\pm0.38}$ & $90.42{\scriptstyle\pm0.38}$ \\
  & mlp-clf (gemma-2-2B)     & $93.7{\scriptstyle\pm6.4}$ & $63{\scriptstyle\pm1}$ & $64{\scriptstyle\pm0}$ & $0.469{\scriptstyle\pm0.006}$ & $-0.18{\scriptstyle\pm0.25}$ & $90.36{\scriptstyle\pm0.25}$ \\
\cdashline{2-8}[2pt/2pt]
  & Linear SAE probe         & $9710$ & -- & -- & $0.157{\scriptstyle\pm0.001}$ & $+18.57{\scriptstyle\pm0.98}$ & $88.25{\scriptstyle\pm0.98}$ \\
  & verb-clf (gemma-3-1B-it) & $30.0{\scriptstyle\pm5.0}$ & $27{\scriptstyle\pm10}$ & $30{\scriptstyle\pm6}$ & $0.094{\scriptstyle\pm0.020}$ & $+0.82{\scriptstyle\pm1.89}$ & $70.50{\scriptstyle\pm1.89}$ \\
\midrule
\multirow{4}{*}{PubMed-20k}
  & Linear SAE probe         & $25893$ & -- & -- & $0.453{\scriptstyle\pm0.000}$ & $-2.47{\scriptstyle\pm0.87}$ & $73.58{\scriptstyle\pm0.87}$ \\
  & mlp-clf (gemma-2-2B)     & $100.0{\scriptstyle\pm2.6}$ & $47{\scriptstyle\pm10}$ & $62{\scriptstyle\pm1}$ & $0.526{\scriptstyle\pm0.004}$ & $-0.02{\scriptstyle\pm0.23}$ & $76.03{\scriptstyle\pm0.23}$ \\
\cdashline{2-8}[2pt/2pt]
  & Linear SAE probe         & $7853$ & -- & -- & $0.087{\scriptstyle\pm0.000}$ & $+43.04{\scriptstyle\pm1.66}$ & $70.92{\scriptstyle\pm1.66}$ \\
  & verb-clf (gemma-3-1B-it) & $23.3{\scriptstyle\pm2.5}$ & $20{\scriptstyle\pm4}$ & $23{\scriptstyle\pm3}$ & $0.146{\scriptstyle\pm0.012}$ & $+15.93{\scriptstyle\pm2.04}$ & $43.81{\scriptstyle\pm2.04}$ \\
\bottomrule
\end{tabular}%
}
\end{table*}

%% file: content/figures/rounds_all_datasets.tex
\begin{figure}[tp]
    \centering
    \includegraphics[width=\textwidth]{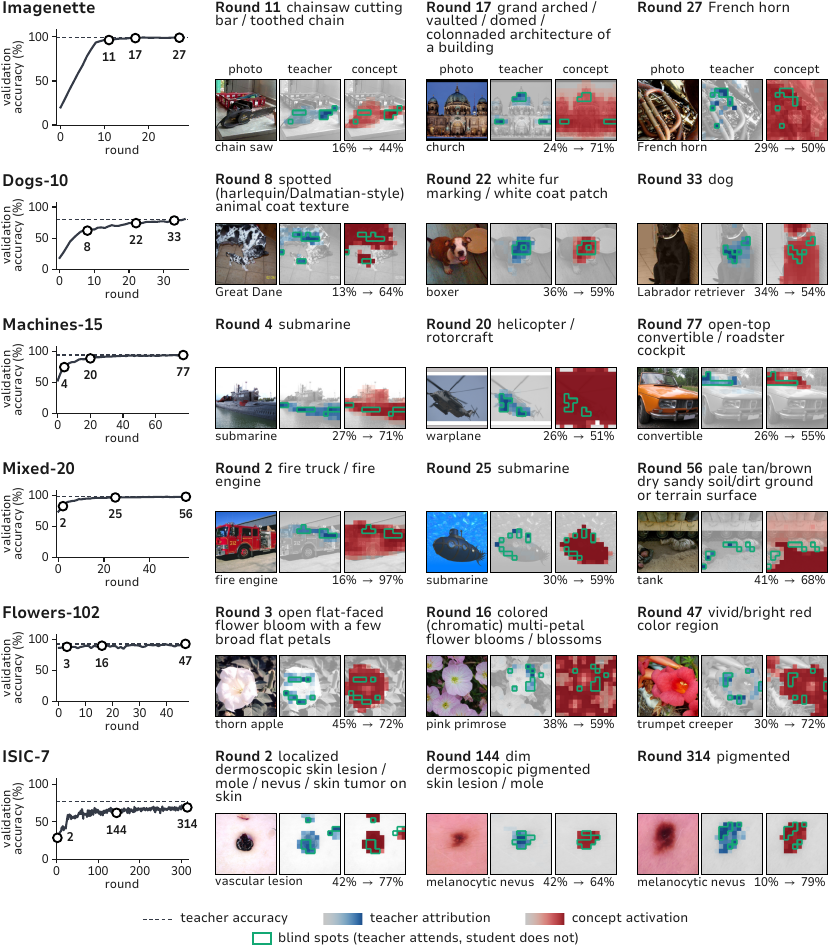}
    \caption{\textbf{Concept selection round by round on all six vision datasets.} For each dataset, we show the validation accuracy of the current predictor $f_\mathcal{S}$ (left) and the concepts and blind spots for three rounds (right). The dashed line is the teacher's validation accuracy. For each round, we show one training image from its focus set on which the added concept fires, with the teacher's attribution and the concept's activation. Green outlines mark the round's blind spots, i.e.\ the tokens the teacher attends to, but $f_\mathcal{S}$ does not. Below each image are its class and the probability $f_\mathcal{S}$ assigns to it before $\to$ after the round. 
    }
    \label{fig:rounds_all}
\end{figure}

%% file: content/figures/adaptive_supp.tex
\begin{figure}[p]
    \centering
    \includegraphics[width=\textwidth]{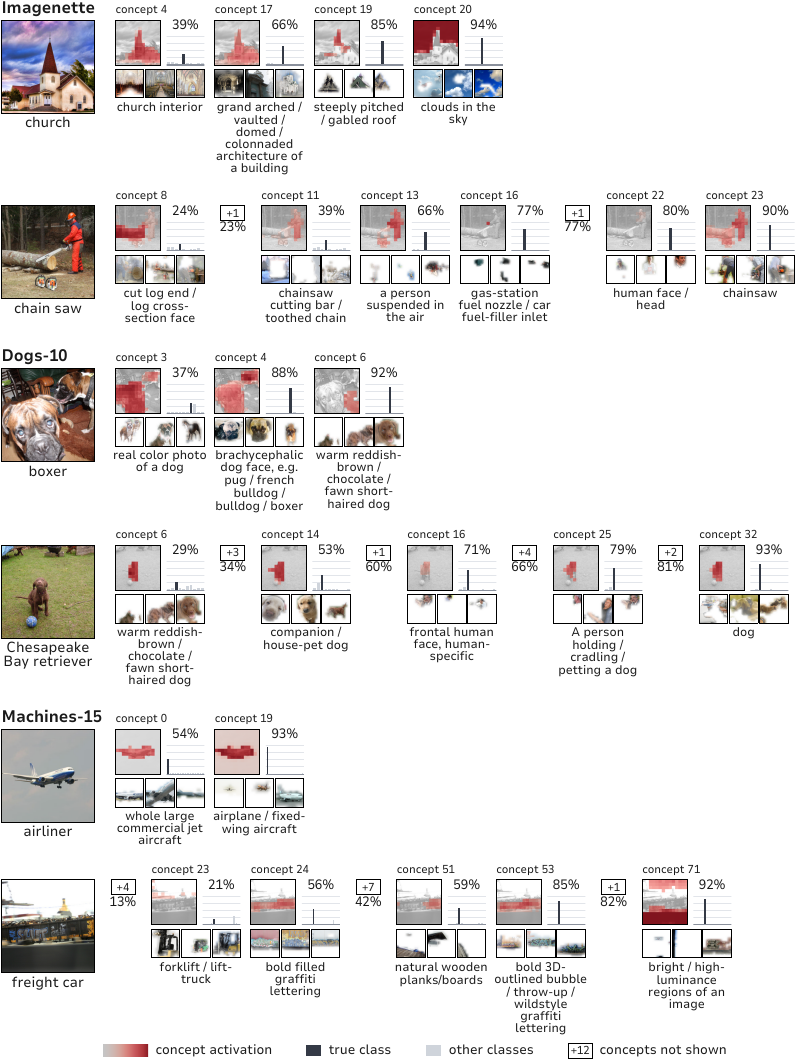}
    \caption{\textbf{Adaptive inference on the other five datasets.} Two test images per dataset, one per row, as \autoref{fig:adaptive_mixed} shows them for Mixed-20. Imagenette, Dogs-10 and Machines-15 here; Flowers-102 and ISIC-7 in \autoref{fig:adaptive_supp_2}.}
    \label{fig:adaptive_supp}
\end{figure}

\begin{figure}[tbp]
    \centering
    \includegraphics[width=\textwidth]{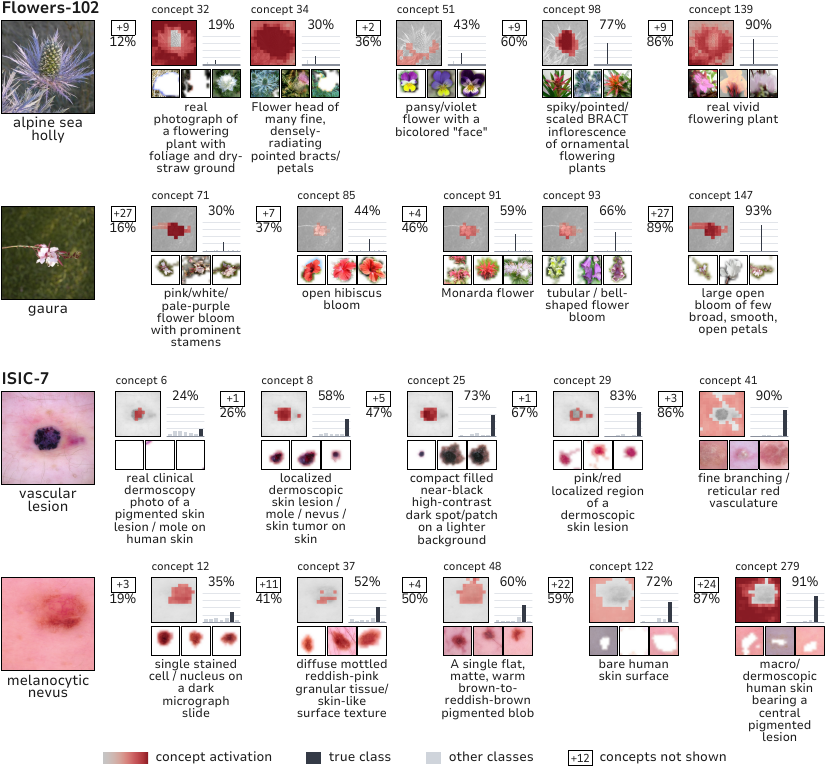}
    \caption{\textbf{Adaptive inference on Flowers-102 and ISIC-7}, continued from \autoref{fig:adaptive_supp}.}
    \label{fig:adaptive_supp_2}
\end{figure}

%% file: content/figures/concept_consistency.tex
\begin{figure}[p]
    \centering
    \includegraphics[width=\textwidth]{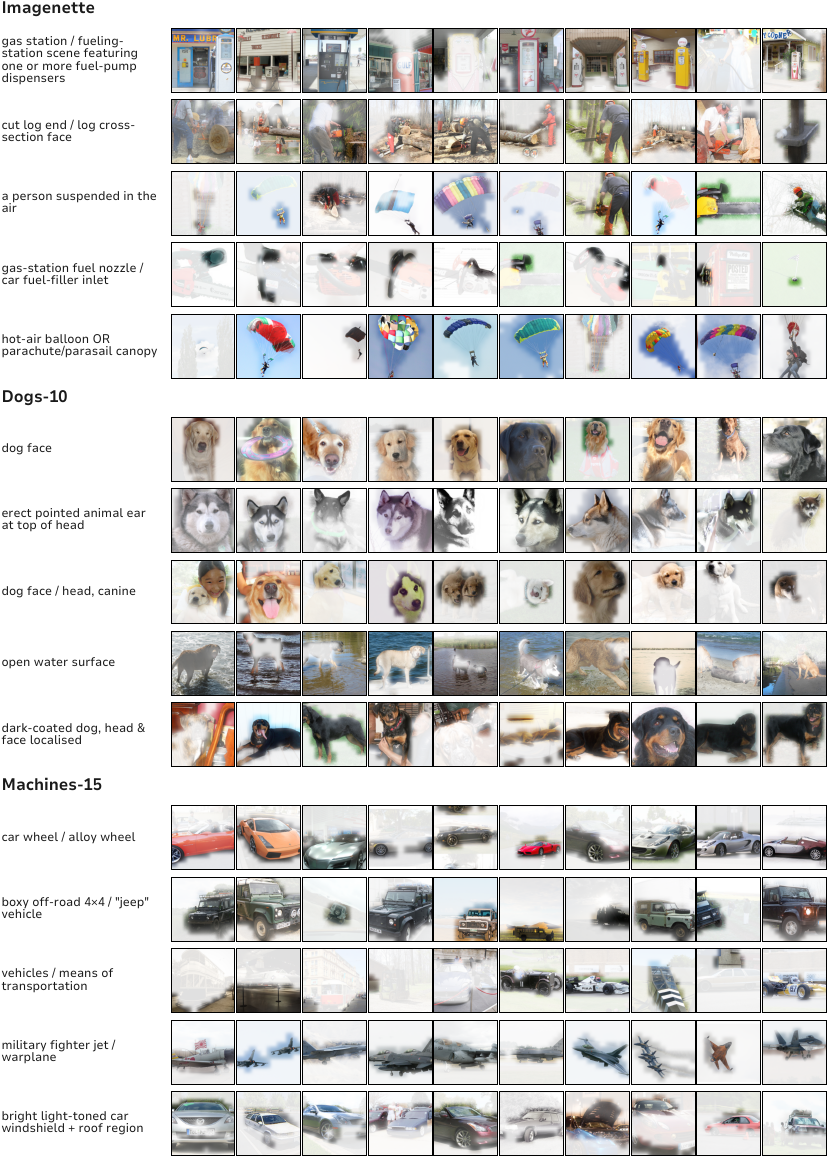}
    \caption{\textbf{Concept examples.} Five concepts per dataset, drawn at random, one per row, each with the ten test images it fires on most strongly. Each image keeps its colour where the concept fires, scaled to its own strongest patch, and fades elsewhere. Imagenette, Dogs-10 and Machines-15 here; Mixed-20, Flowers-102 and ISIC-7 in \autoref{fig:consistency_2}.}
    \label{fig:consistency}
\end{figure}

\begin{figure}[p]
    \centering
    \includegraphics[width=\textwidth]{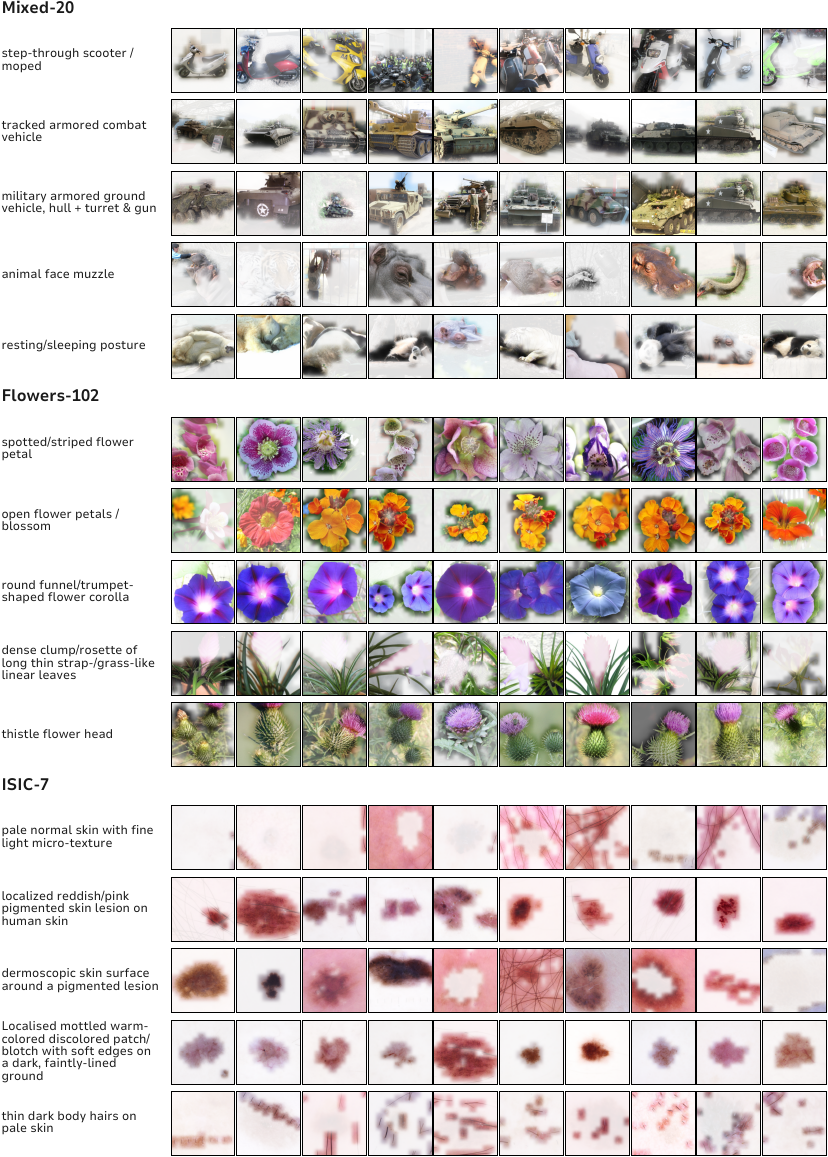}
    \caption{\textbf{Concept examples}, continued from \autoref{fig:consistency}: Mixed-20, Flowers-102 and ISIC-7.}
    \label{fig:consistency_2}
\end{figure}

%% file: content/tables/imagenet_subsets.tex
\begin{table}[t]                                                               
  \centering                                                                   
  \small                                                                       
  \caption{ImageNet-1k classes used in our three curated subsets.}             
  \label{tab:imagenet_subsets}                                                 
  \begin{tabular}{@{}lp{0.72\linewidth}@{}}                                    
  \toprule                                                                     
  Subset & Classes \\                                                          
  \midrule                                                                     
  Dogs (10) & Golden retriever, Labrador retriever, Chesapeake Bay retriever,  
    German shepherd, Rottweiler, Doberman, Boxer, Great Dane, Siberian husky,  
    Eskimo dog \\                                                         
  \addlinespace                                                           
  Machines (15) & Sports car, Station wagon, Convertible, Jeep, Pickup,   
    Garbage truck, Trailer truck, Steam locomotive, Streetcar, Freight car,   
    Airliner, Warplane, Speedboat, Submarine, Forklift \\             
  \addlinespace                                                                    
  Mixed (20) & \emph{Animals:} Lion, Tiger, African elephant, Zebra, Giant panda,  
    Polar bear, Gorilla, Flamingo, Ostrich, Hippopotamus;  
    \emph{Machines:} Sports car, Warplane, Submarine, Steam locomotive, Forklift, 
    Ambulance, Tank, Space shuttle, Fire engine, Motor scooter \\                        
  \bottomrule                        
  \end{tabular} 
\end{table}    

%% file: content/tables/verbalizer.tex
\begin{table}[h]                                  
  \centering                                      
  \small  
  \caption{Verbalizer prompts (wrapped in the Gemma-3 chat template) and class-to-token mappings.}           
  \label{tab:verbalizer}                        
  \resizebox{\textwidth}{!}{%
  \begin{tabular}{l p{0.52\linewidth} l}          
  \toprule                                        
  Dataset & Prompt & Class $\rightarrow$ token \\                  
  \midrule    
  AG News & \texttt{Article: "\{text\}"} \newline \texttt{Which topic best fits this article: World, Sports, Business, or Technology? Answer with exactly one word.}                 
  & World, Sports, Business, Sci/Tech $\rightarrow$ Technology \\ 
  PubMed 20k RCT & \texttt{Sentence from a medical research abstract: "\{text\}"} \newline \texttt{Does this sentence state the study's background, objective, methods, results, or  
  conclusion? Answer with exactly one word.}   
  & Background, Objective, Methods, Results, Conclusion \\             
  \bottomrule                  
  \end{tabular} 
  }
\end{table}  

%% file: content/tables/hyperparams.tex
\begin{table}[h]
  \centering
  \caption{Hyperparameters used throughout \method's selection and training pipeline. Where vision and text differ, both values are given; a single value applies to both. For text, G2 = Gemma-2-2B MLP-teacher runs, G3 = Gemma-3-1B-it verbalizer runs. Accuracy thresholds are in percentage points.}
  \label{tab:hyperparams}
  \resizebox{\linewidth}{!}{%
  \begin{tabular}{@{}llll@{}}
  \toprule
  \textbf{Stage} & \textbf{Parameter} & \textbf{Vision} & \textbf{Text} \\
  \midrule
  \multirow{5}{*}{Backbone / Teacher}
   & Teacher backbone & CLIP-DINOiser (ViT-B/16), CLS & G2: Gemma-2-2B, last-token resid.\ (layer 19) / G3: Gemma-3-1B-it (frozen verbalizer) \\
   
   & Teacher head & \multicolumn{2}{p{12cm}}{\centering MLP, hidden dim 512, dropout 0.1 (G3: none, frozen LM)} \\ 
   & Teacher optimizer & \multicolumn{2}{p{12cm}}{\centering Adam, lr $10^{-3}$, weight decay $10^{-4}$ (G3: n/a)} \\
   & Teacher batch size / epochs & 256 / 40 & full-batch / 60 (early stop, patience 12) \\
   & Teacher selection metric & \multicolumn{2}{p{12cm}}{\centering balanced val.\ accuracy} \\
  \midrule
  \multirow{4}{*}{SAE}
   & Architecture & Matryoshka BatchTopK & G2: TopK (SAEBench) / G3: JumpReLU (Gemma Scope 2) \\
   & Dictionary size $M$ & 8192 & G2: 64{,}512 / G3: 16{,}384 \\
   & Sparsity & $k=12$ & G2: $k=40$ / G3: target $L_0\approx60$ \\
   & Pretraining corpus & CC12M & G2: The Pile (~400M tokens) / G3: Gemma~3 pretraining distribution + chat data (oasst1, LMSYS-Chat-1M) \\
  \midrule
  \multirow{6}{*}{SAE Fine-tuning}
   & Objective & \texttt{teacher\_kl\_recon} & \texttt{teacher\_kl} \\
   & Fine-tuning teacher & patch-attention MLP (frozen)\textsuperscript{b} & same teacher as discovery \\
   & KL temperature / recon.\ weight & 1.0 / 0.1 & 1.0 / 0 \\
   & Optimizer & \multicolumn{2}{p{12cm}}{\centering Adam, lr $10^{-4}$} \\
   & Batch size / max epochs & 64 / 40 (ISIC: 20) & 32 / 20 \\
   & Early stopping & val.\ CE increase, patience 5, 200 warmup steps & best val.\ loss epoch \\
  \midrule
  \multirow{2}{*}{Rescaling}
   & Percentile $p$ & \multicolumn{2}{p{12cm}}{\centering 99} \\
   & Clip value $\tau_\text{max}$, $\epsilon$ & \multicolumn{2}{p{12cm}}{\centering $1.0$, $10^{-6}$} \\
  \midrule
  \multirow{7}{*}{Candidate Proposal}
   & Focus set size $k$ & 200 & G2: 800 (AG News) / 500 (PubMed); G3: 500 \\
   & Blind-spot token count $b$ & 10 & G2: 6 (AG News) / 10 (PubMed); G3: 10 \\
   & Per-sample top candidates $r$ & 5 (10 for Flowers, ISIC) & 30 \\
   & Min distinct-sample count $d_\text{min}$ & \multicolumn{2}{p{12cm}}{\centering 3} \\
   & Shortlist size $q$ & \multicolumn{2}{p{12cm}}{\centering 30} \\
   & Activation threshold $\theta$ & \multicolumn{2}{p{12cm}}{\centering 0.01} \\
   & Redundancy filter (cosine threshold) & disabled & $0.6$ \\
    & Class-focused jump-start round & enabled: Machines-15, Mixed-20, Flowers-102, ISIC-7 & -- \\
    \midrule
  \multirow{2}{*}{Candidate Scoring}
   & Scoring $\epsilon$ (Eq.~\ref{eq:scoring}) & \multicolumn{2}{p{12cm}}{\centering $10^{-8}$} \\
   & Minimum score threshold $s_\text{min}$ & $10^{-3}$ (immediate stop) & $10^{-3}$ (3 consecutive rounds) \\
  \midrule
  \multirow{7}{*}{Retraining / Stopping}
   & Student optimizer & \multicolumn{2}{p{12cm}}{\centering lr $10^{-2}$, weight decay $10^{-4}$} \\
   & Student batch / epochs per round & 128 / 5 (64 / 3--10)\textsuperscript{a} & 32 / 5 \\
   & Accuracy gap tolerance $\delta_\text{acc}$ & 0.1 & 0.0 \\
   & Validation plateau margin $\eta$ & 0.01 & 0.1 \\
   & EV plateau margin $\eta_\text{EV}$ & \multicolumn{2}{p{12cm}}{\centering 0.001} \\
   & Patience $P$ & 10 / 20\textsuperscript{a} & 30 \\
   & Max rounds $T_\text{max}$ & 300 / 500\textsuperscript{a} & 200 \\
  \midrule
  \multirow{3}{*}{Nested Dropout}
   & Prefix sampling & \multicolumn{2}{p{12cm}}{\centering uniform over $\{1,2,4,\dots,2^i\}\cup\{|\mathcal{S}^\star|\}$} \\
   & Epochs / lr / patience & \multicolumn{2}{p{12cm}}{\centering 100 / $10^{-3}$ / 20} \\
  \midrule
  Evaluation & Seeds & \multicolumn{2}{p{12cm}}{\centering 42, 123, 7 (data split fixed, seed 42)} \\
  \bottomrule
  \end{tabular}
  }
\end{table}

%% file: content/prompts/maia_prompts.tex
\lstdefinestyle{maiaprompt}{
  basicstyle=\ttfamily\scriptsize,
  columns=fullflexible,
  keepspaces=true,
  breaklines=true,
  breakatwhitespace=false,
  frame=single,
  framesep=4pt,
  xleftmargin=4pt,
  xrightmargin=4pt,
  showstringspaces=false,
}

\paragraph{MAIA prompts.}
After 25 rounds the MAIA agent is prompted to conclude (Listing~\ref{lst:maia-final}).
The agent receives the system prompt (Listing~\ref{lst:maia-system}), the user prompt (Listing~\ref{lst:maia-user}), and the exemplar evidence (Listing~\ref{lst:maia-evidence}), in that order.

\begin{lstlisting}[style=maiaprompt,caption={MAIA system prompt (compact variant for 32k-context local VLMs).},label={lst:maia-system}]
You are MAIA, an interpretability agent labeling one visual concept/neuron in an AnyBottle SAE/CBM run.

You already have initialized objects named `system` and `tools`. Never import, construct, overwrite, or reassign
`System`, `Tools`, `system`, or `tools`.

Available calls:
- `tools.dataset_exemplars(system)` returns top activating real exemplars as `(activation, image_b64)` pairs.
- `tools.text2image(list[str])` generates synthetic images from text prompts.
- `tools.edit_images(base_images: list[str], editing_prompts: list[str])` edits given base images/prompts.
- `system.call_neuron(images)` returns `(activation_list, activation_map_list)` for images.
- `tools.display(...)` records images/text/results for you to inspect after execution.

Use only existing `system` and `tools`. Write experiments as Python code blocks. Prefer small experiments:
1-3 prompts or edits per code block. After seeing results, update hypotheses. When confident, stop writing code and
return final output exactly as:
[DESCRIPTION]: <specific visual concept>
[LABEL 1]: <short label>
[LABEL 2]: <optional second label>
[CONFIDENCE]: <0-100> - <one short reason, based on the evidence you gathered>
\end{lstlisting}

\begin{lstlisting}[style=maiaprompt,caption={MAIA user prompt.},label={lst:maia-user}]
Your task is to identify the visual concept that maximally activates the current neuron/concept.

You are given ground-truth top activating exemplar evidence in the conversation. Use it as primary evidence.
You may run small synthetic `tools.text2image` or `tools.edit_images` experiments to test hypotheses, then call
`system.call_neuron` and `tools.display` to view activations/results.

Rules:
- Do not use examples from instructions as evidence.
- Do not reinitialize `system` or `tools`.
- Do not write a final description in the same response as code. If you write code, wait for execution results.
- Keep generated prompts specific and very few.

Begin by proposing one small experiment, or provide the final answer if the exemplar evidence is already decisive.

Important for this AnyBottle run: API examples in the system prompt (for example dogs, cats, grass, landscapes, or generated images) are usage examples only. They are not evidence about the current neuron. Never use those examples in the final label unless they are visibly present in the actual attached top activating exemplars.

The objects named `system` and `tools` are already initialized for the current AnyBottle SAE concept. Do not import, construct, overwrite, or reassign `System`, `Tools`, `system`, or `tools`. In code blocks, call methods on the existing objects, for example `tools.text2image(...)`, `tools.edit_images(...)`, `system.call_neuron(...)`, and `tools.dataset_exemplars(system)`.

Localisation: `system.call_neuron(images)` returns, per image, the concept's activation and a turbo heatmap of where in the image it responds. For a closer look use `overlays, info = system.concept_activation_map(images)` -- `info[i]` has a small `grid` (g x g ints 0-100, row 0 = top of the image) and a `summary` with `peak_xy` (x,y image fractions), `active_fraction`, `top10pct_mass` (near 1.0 = a sharp point, ~0.1 = spread evenly) and `quadrant_mass` [top-left, top-right, bottom-left, bottom-right]. Use it to tell a localised object/part (peak, low active_fraction, high top10pct_mass) from a whole-scene or texture concept (spread out). Reflect this in the label, e.g. 'dog's face (localised)' vs 'outdoor grass texture (diffuse)'.

Confidence: whatever final-answer format the instructions above specify, add one more line at the end -- exactly:
[CONFIDENCE]: <0-100> - <one sentence on what would raise or lower it, e.g. how many exemplars fit vs. don't, whether experiments causally confirmed the hypothesis or you only had exemplar evidence, conflicting concepts in the top exemplars>. Base it on the evidence you actually gathered, not general confidence in the task.

Attribute decomposition: an object-level label (e.g. 'dog') is a starting hypothesis, not a final answer. Before finalizing, decompose it along these dimensions and identify which one actually drives the activation:
  - shape/geometry (e.g. the silhouette or outline, independent of what fills it)
  - texture/material (e.g. fur, fabric weave, metal sheen, independent of the object's identity)
  - color (does the concept survive recoloring/desaturation, or does it collapse?)
  - part vs. whole (is it the whole object, or one localised part, e.g. ears, snout, a specific edge?)
  - context/background (does it need the surrounding scene, or does it fire on the object/part alone, cropped out?)
If `tools.edit_images`/`tools.text2image` are available, use them to test these directly: e.g. recolor or desaturate an exemplar and re-run `system.call_neuron` to see if activation survives; swap or blur the texture while keeping the same silhouette; crop to just the candidate part vs. the full object vs. just the background. An attribute the concept does NOT need should be explicitly ruled out in your hypothesis list, not just omitted. If those tools are unavailable in this run, still push the label as far down this hierarchy as the exemplar evidence alone supports (e.g. note consistent fur texture across varied dog breeds/poses as evidence for texture over shape), and say in [CONFIDENCE] that this wasn't causally tested.
\end{lstlisting}

\begin{lstlisting}[style=maiaprompt,caption={Exemplar-evidence message (template; braces are filled in per concept).},label={lst:maia-evidence}]
Ground-truth visual evidence for the current neuron/concept follows. These are the actual mined exemplars for this run; API examples in the system prompt are not evidence and must not be used for the final label.

Top activating exemplar activations: {a_1, ..., a_4}

Top activating exemplar summary from the image-description model:
{summary of the top-4 exemplars}

For this local-VLM run, the raw exemplar image payloads were used to produce the summary above but are not repeated in this reasoning prompt because they exceed the model context window. Base your final description and labels on this visual summary and any subsequent displayed experiment results.
\end{lstlisting}

\begin{lstlisting}[style=maiaprompt,caption={Final-answer prompt, issued after 25 rounds without a final answer (from OpenMAIA).},label={lst:maia-final}]
Please provide you final description in the following format:
[DESCRIPTION]: <final description> ## Your description should be selective (e.g. very specific: "dogs running on the grass" and not just "dog") and complete (e.g. include all relevant aspects the neuron is selective for). In cases where the neuron is selective for more than one concept, include in your description a list of all the concepts separated by logical "OR".
[LABEL]: <final label drived from the hypothesis or hypotheses> ## a label for the neuron generated from the hypothesis (or hypotheses) you are most confident in after running all experiments. They should be concise and complete, for example, "grass surrounding animals", "curved rims of cylindrical objects", "text displayed on computer screens", "the blue sky background behind a bridge", and "wheels on cars" are all appropriate. You should capture the concept(s) the neuron is selective for. Only list multiple hypotheses if the neuron is selective for multiple distinct concepts. List your hypotheses in the format:
[LABEL 1]: <label 1>
[LABEL 2]: <label 2>
\end{lstlisting}